\documentclass[10pt,twocolumn,letterpaper]{article}

\usepackage[final]{iccv}      

\usepackage{xcolor} 
\usepackage{multirow}
\usepackage{rotating}
\usepackage{xcolor}
\usepackage{colortbl}
\usepackage{booktabs}
\usepackage{algorithmic}
\usepackage{algorithm}
\usepackage{arydshln}
\usepackage{cuted} 

\usepackage{listings}
\usepackage{xcolor}

\usepackage[dvipsnames]{xcolor}

\definecolor{iccvblue}{rgb}{0.21,0.49,0.74}
\usepackage[pagebackref,breaklinks,colorlinks,allcolors=iccvblue]{hyperref}

\def\paperID{12454} 
\def\confName{ICCV}
\def\confYear{2025}

\title{Co-STAR: Collaborative Curriculum Self-Training with Adaptive Regularization for Source-Free Video Domain Adaptation}

\author{Amirhossein Dadashzadeh, Parsa Esmati, and Majid Mirmehdi\\
University of Bristol, UK\\
{\tt\small \{a.dadashzadeh, parsa.esmati, m.mirmehdi\}@bristol.ac.uk}
}

\begin{document}
\maketitle
\begin{abstract}

Recent advances in {Source-Free Unsupervised Video Domain Adaptation (SFUVDA) leverage vision-language models to enhance pseudo-label generation. However, challenges {such as} noisy pseudo-labels and over-confident predictions limit their effectiveness in adapting {well} across domains. We propose Co-STAR, a novel framework that integrates curriculum learning with collaborative self-training between a source-trained teacher and a contrastive vision-language  model (CLIP). Our curriculum learning approach employs a reliability-based weight function that measures bidirectional prediction alignment between the teacher and CLIP, balancing between confident and uncertain predictions. This function preserves uncertainty for difficult samples, while prioritizing reliable pseudo-labels when the predictions from both models closely align. To further improve adaptation, we propose Adaptive Curriculum Regularization, which modifies the learning priority of samples in a probabilistic, adaptive manner based on their confidence scores and prediction stability, mitigating overfitting to noisy and over-confident samples. Extensive experiments across multiple video domain adaptation benchmarks demonstrate that Co-STAR consistently outperforms state-of-the-art {SFUVDA} methods. Code is available at: https://github.com/Plrbear/Co-Star}
\end{abstract}

\section{Introduction}
\label{sec:intro}

\begin{figure}[t]
\centerline{\includegraphics[scale=0.75]{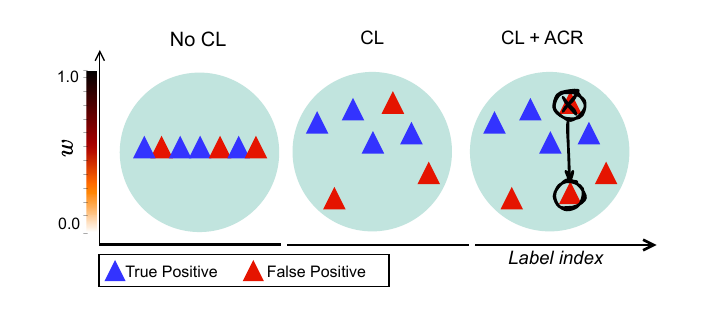}}
\caption{{\bf Impact of curriculum learning (CL) and Adaptive Curriculum Regularization (ACR) on data reliability.}  Without CL, all data points are treated equally regardless of reliability. With CL, reliable data points (blue triangles/likely true positives) are prioritized, while less reliable points (red triangles/potential false positives) receive less influence. With CL+ACR, our method further refines the process by identifying and adjusting overconfident but incorrect predictions, preventing the model from reinforcing errors during training. Here, $w$ is a weight value that determines how much the model trusts each sample's pseudo-label during training.}
\label{Fig:fig-intro} 
\end{figure}

Recent advances in deep learning have significantly improved action recognition, with models like CNNs \cite{feichtenhofer2016convolutional, feichtenhofer2017spatiotemporal}
 and vision transformers \cite{han2022survey} now able to capture complex patterns in video data. This progress has driven applications in fields such as healthcare \cite{htet2022hmm}, intelligent human–machine
interfaces \cite{presti20163d}, and identity recognition \cite{paul2014survey}. However, these models face a {fundamental} challenge when tested on new, unseen data distributions where their performance often declines due to domain shifts. In real-world scenarios, where labelling data for each new domain is impractical, {unsupervised video domain adaptation (UVDA) becomes viable} \cite{sahoo2021contrast, da2022dual, da2022unsupervised, reddy2024unsupervised}. 
UVDA can address this challenge by enabling models to transfer knowledge from labelled source data to unlabelled target data, but accessing source data is not always possible due to {privacy, legal or other} restrictions \cite{ishikawa2024learnable}, leading to  source-free unsupervised video domain adaptation (SFUVDA) \cite{ATCON,STHC, EXTERN, zara2023unreasonable}. SFUVDA relies solely on a pretrained source model and unlabeled target videos to provide an unsupervised solution.


The use of vision-language models like CLIP \cite{radford2021learning} has recently advanced source-free video domain adaptation \cite{zara2023unreasonable}. DALL-V \cite{zara2023unreasonable} leverages CLIP’s robust, domain-agnostic representations for pseudo-labeling, significantly enhancing SFUVDA's effectiveness. Despite improved performance, CLIP-generated pseudo-labels can still introduce {noise, which may arise from overconfident, yet incorrect, predictions,} or from predictions with low confidence, indicating uncertainty. Both types can mislead the model, causing errors or unstable learning, particularly during early adaptation when reliable feature extraction is vital \cite{roy2021curriculum, wu2020curricula, CSFDA}.

To address these issues, we propose Co-STAR (Collaborative Curriculum Self-Training with Adaptive Regularisation),
a novel framework that introduces curriculum learning into collaborative self-training for source-free video domain adaptation. Through the dynamic weighting of pseudo-labels generated collaboratively by a pretrained teacher model and CLIP, Co-STAR guides the adaptation process from reliable to challenging samples. 

A central component of Co-STAR is Adaptive Curriculum Regularization (ACR), which enhances curriculum learning by adaptively and probabilistically adjusting the importance of training samples. ACR examines both how confident the model is about a prediction and how stable this prediction remains over time. As training progresses, ACR's influence gradually increases, helping prevent the model from relying too heavily on overconfident pseudo-labels that may be incorrect, while preserving the benefits of curriculum learning.
{Fig. \ref{Fig:fig-intro} illustrates the impact of curriculum learning and ACR on the importance of unlabelled target data points, which emphasizes how Co-STAR adapts training to prioritize reliable samples and adjust overconfident pseudo-labels.}

Our contributions in this work are:
\begin{itemize}
\item {We propose a novel source-free {unsupervised} video domain adaptation framework, Co-STAR, that integrates curriculum learning into collaborative self-training between a source-trained teacher and {the} CLIP model, where a reliability-based weighting function dynamically balances between uncertain and confident predictions.}

\item {{We introduce adaptive curriculum regularization, a probabilistic mechanism that adaptively adjusts curriculum weights to avoid overfitting on noisy, overconfident pseudo-labels.}}


\item {We perform comprehensive experiments across three video domain adaptation benchmarks, and show that Co-STAR consistently outperforms existing SFUVDA methods through its collaborative curriculum learning approach. Our detailed ablation studies validate how each component - from CLIP integration to adaptive regularization - contributes to the overall performance gains.}
\end{itemize}

\section{Related Work}
\label{sec:Related works}
We review the latest works most relevant to our research, focusing on recent advances in Source-Free Unsupervised Video Domain Adaptation (SFUVDA) and curriculum learning strategies for unsupervised domain adaptation (UDA).

\noindent {\bf{SFUVDA.}}
To address the challenges of source-free unsupervised video domain adaptation, a majority of recent studies have focused on adapting self-supervised techniques \cite{ATCON,EXTERN,STHC, zara2023unreasonable}. ATCoN \cite{ATCON}, for instance, employs both feature and source prediction temporal consistency to help stabilize the model’s predictions and reduce sensitivity to temporal variations. EXTERN \cite{EXTERN} improves prediction stability by adopting a mask-to-mix technique, combining selected unmasked clips to align with the entire sequence’s features. Through endo- and exo-temporal regularizations, EXTERN enhances feature discriminability and consistency across temporal clips. STHC \cite{STHC} improves adaptation by introducing spatial consistency alongside temporal consistency, allowing the model to achieve stable predictions across spatially augmented frames. To further reinforce prediction stability, STHC incorporates historical consistency via a memory bank that stores prior predictions to ensure coherence over time. 

Following recent trends in using vision-language models (VLMs) for action recognition tasks \cite{xu2021videoclip, wang2021actionclip,wu2023revisiting}, DALL-V \cite{zara2023unreasonable} leverages the zero-shot capabilities of CLIP \cite{radford2021learning} to address domain gaps in video data. By performing pseudo-labeling with CLIP and refining these labels through an ensemble distillation process, DALL-V achieves significant performance improvements, outperforming previous state-of-the-art methods in SFUVDA. While DALL-V achieves impressive results, its use of CLIP-generated pseudo-labels can still introduce noise. This integration often leads to overconfident or unstable predictions, particularly during early adaptation. We deal with these challenges byintroducing curriculum learning through Co-STAR to have a more stable and reliable adaptation process across domains.

\noindent {\bf{Curriculum Learning for UDA.}} Curriculum learning for UDA involves an adaptive, progressive strategy, gradually refining the model’s ability to address domain shifts by starting with simpler tasks and moving to more complex ones \cite{lian2019constructing, sakaridis2019guided, zhang2019curriculum, choi2019pseudo, roy2021curriculum, chen2021unsupervised, CSFDA}. For instance, Zhang et al. \cite{zhang2019curriculum} start with global label distributions and move toward pixel-level tasks, facilitating adaptation from synthetic to real-world urban scenes. 
Recently, Karim et al. \cite{CSFDA} extended curriculum learning to source-free domain adaptation with C-SFDA, {demonstrating that simple confidence thresholding or loss-based pseudo-label selection methods \cite{dasgupta2022overcoming} are inadequate in SFUDA settings.} This approach categorizes target samples as reliable or unreliable based on confidence and uncertainty, focusing first on reliable samples and then using them to refine information for unreliable ones. C-SFDA achieves state-of-the-art results across image classification and semantic segmentation. 
However, this approach does not fully address large domain gaps, as stated in~\cite{CSFDA}. When the gap is substantial, most target data may be considered as unreliable which can limit effective learning. In contrast, we propose a collaborative curriculum learning approach that avoids binary reliability labels by leveraging both a domain-agnostic vision-language model {(CLIP) \cite{radford2021learning}} and a domain-specific pretrained teacher (and we extend this to video). This collaboration provides more precise reliability scores, allowing each data point to contribute meaningfully to adaptation, even under significant domain shifts.

\begin{figure*}[t]
\centerline{\includegraphics[scale=0.81]{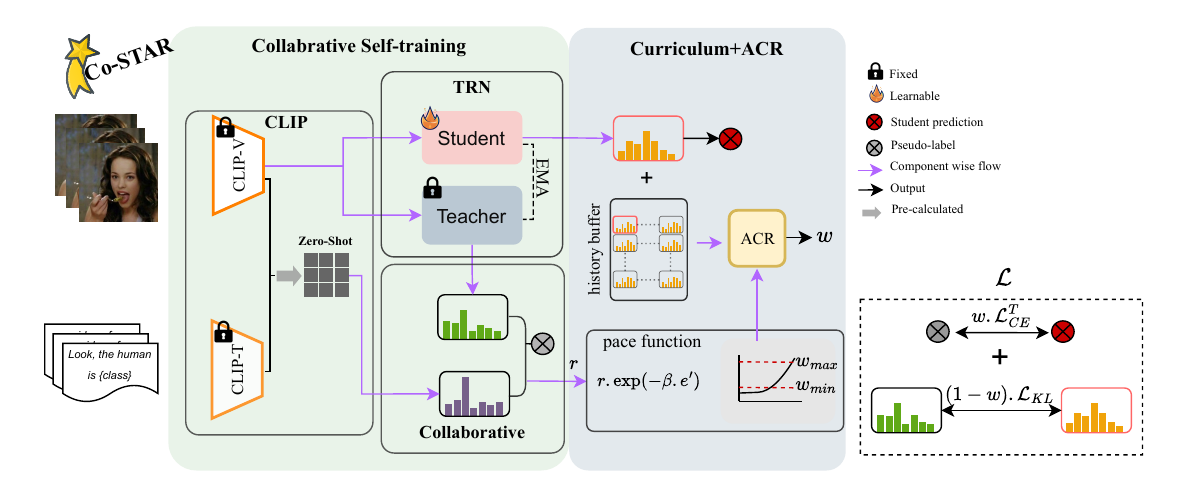}}
\caption{{\textbf{Overview of the Co-STAR framework.} The collaborative self-training component leverages a teacher-student architecture, where both the CLIP and teacher models contribute to pseudo-label generation. The curriculum learning component dynamically adjusts the balance between CE and KL divergence losses based on reliability scores \( r \). The ACR module refines this balance by monitoring prediction stability through a {history buffer}. The student model is then trained using this adaptive balance, effectively combining knowledge distillation with pseudo-label learning for robust adaptation. {Note that CLIP's zero-shot predictions are pre-computed before training to reduce computational overhead.}}}

\label{Fig:fig-method} 
\end{figure*}

\section{Method}


\noindent{\bf{Problem definition.}}


In SFUVDA, the objective is to adapt a model trained on a labeled source domain to perform well on an unlabeled target domain, without access to the original source data during adaptation. Let the source dataset be \( \mathcal{D}^S = \{ (x_i^S, y_i^S) \}_{i=1}^{X^S} \), where \( x_i^S \) are video sequences, \( y_i^S \in \{1, 2, ..., K\} \) represent action labels for \( K \) classes, and \(X^S\) is the number of samples in the source domain. Let \( F_{\text{teach}}(\cdot) = \mathcal{R}_{\text{teach}}(G_v(\cdot)) \) denote our initial model architecture, where $G_v$ is a frame-level visual encoder and  $\mathcal{R}_{\text{teach}}:\mathbb{R}^{d \times n}\rightarrow\mathbb{R}^d$, is a Temporal Relation Network (TRN)~\cite{zhou2018temporal},  $d$ is the feature dimensionality and $n$ is the number of frames. The model is trained on $\mathcal{D}^S$ to obtain source domain knowledge.

The target domain is represented by an unlabeled dataset \( \mathcal{D}^T = \{ x_i^T \}_{i=1}^{X^{T}} \), containing \( X^{T} \) video sequences with no associated labels. Although both domains share the same label space, the distribution gap between them leads to a domain shift, which reduces the source model's performance on \( \mathcal{D}^T \). The challenge is to learn \( F_{\text{stud}}(\cdot) = \mathcal{R}_{\text{stud}}(G_v(\cdot)) \) to accurately predict target labels \(y_i^T\) without labels being available or access to \( \mathcal{D}^S \).

\subsection{Co-STAR}
\label{Co-STAR}
{The proposed Co-STAR framework has two major components operating concurrently to enable source-free unsupervised video domain adaptation.} {The first component introduces collaborative self-training, where pseudo-labels are generated using both a source-trained teacher model and a vision-language model (CLIP). The second component incorporates curriculum learning on the target domain to prioritize samples based on their reliability, with Adaptive Curriculum Regularization progressively controlling overconfidence and enhancing adaptation. The overall Co-STAR framework and its interactions, is shown in Fig. \ref{Fig:fig-method}. Next, we elaborate on the various elements of Co-STAR.}

\vspace*{2mm}
\noindent {\bf{Source Fine-Tuning.}} {To fine-tune their source model, previous SFVDA works~\cite{ATCON, STHC} incorporated a Temporal Relation Network~\cite{zhou2018temporal} on top of their frame-level visual encoder. Inspired by~\cite{zara2023unreasonable}, we initialize our visual encoder ($G_v$) with CLIP's pretrained weights, and freeze the weights to prevent overfitting and then extract spatial features for each frame. However, unlike ~\cite{zara2023unreasonable}, but similar to previous SFVDA works, these frame-level features are then processed by our TRN, $\mathcal{R}_{\text{teach}}$, which combines these features to capture temporal relationships across frames, resulting in our source model $F_{\text{teach}}(\cdot) = \mathcal{R}_{\text{teach}}(G_v(\cdot))$.}
{To train $F_{\text{teach}}$ on the source domain, we apply a cross-entropy (CE) loss:}
\begin{equation}
  \mathcal{L}_{CE}^{S} = \mathbb{E}_{(x_i^S, y_i^S) \sim \mathcal{D}^S} \left[ - \log p(y_i^S | x_i^S) \right] ~,
\end{equation}
where \( p(y_i^S | x_i^S) \) denotes the predicted probability of the correct action label for the video sequence.

\vspace*{2mm}

\noindent {\bf{Collaborative Self-training.}}
We employ a self-training approach using a teacher-student framework, as illustrated in Fig. \ref{Fig:fig-method}. This setup, commonly used in self-training for unsupervised domain adaptation \cite{CSFDA,fang2024source}, consists of our source model $F_{\text{teach}}$ {with weights \(\theta_{\text{teach}}\)} serving as the teacher and a target model $F_{\text{stud}}(\cdot) = \mathcal{R}_{\text{stud}}(G_v(\cdot))$ with weights $\theta_{\text{stud}}$ serving as the student. {Both models share identical network architectures, with the student initialized from the pretrained teacher weights.} The target model learns on the dataset \( \mathcal{D}^{T} \), adapting to the target domain using pseudo-labels.

{We leverage CLIP's zero-shot capabilities to generate pseudo-labels for target domain samples in $\mathcal{D}^T$, simplifying adaptation through collaborative pseudo-labeling between the source-trained teacher and CLIP. Unlike DALL-V \cite{zara2023unreasonable}, our approach does not require ensemble distillation, instead using the MatchOrConf scheme \cite{zhang2023rethinking} to integrate outputs based on confidence and alignment, adopting shared predictions or selecting the more confident model's output.}
We use the generated pseudo-labels \( \tilde{y} \) to train the student model on the target domain with a CE loss,
\begin{equation}
    \mathcal{L}_{CE}^{T} = \mathbb{E}_{x_i^T \sim \mathcal{D}^T} \left[ - \log p(\tilde{y}_i | x_i^T) \right] ~.
\end{equation}
We also use a KL loss $(\mathcal{L}_{\text{KL}})$ to help the student learn from the teacher's knowledge and prevent it from drifting, which provides stability and mitigates overfitting to noisy pseudo-labels by minimizing the Kullback-Leibler (KL) divergence between their softened output distributions, such that 
\begin{equation} \label{eq:kl}
    \mathcal{L}_{KL} = \mathbb{E}_{x_i^T \sim \mathcal{D}^T} \left[ \text{KL}\left( p_{\theta_{\text{stud}}}(x_i^T / \tau) \parallel p_{\theta_{\text{teach}}}(x_i^T / \tau) \right) \right] ~,
\end{equation}
where \( \tau \) is a temperature parameter applied to smooth the outputs, and \( p_{\theta_{\text{stud}}}(x_i^T / \tau) \) and \( p_{\theta_{\text{teach}}}(x_i^T / \tau) \) represent the softened probabilities of the student and teacher for target sample \( x_i^T \).
A weighted sum of these components, with \( w \) controlling the balance between the two losses, provides the total training loss as
\begin{equation}
    \mathcal{L} = (1 - w) \mathcal{L}_{KL} + w \mathcal{L}_{CE}^{T} ~.   \label {eq:finalLoss}
\end{equation}
During self-training, the teacher model's weights \( \theta_{\text{teach}} \) remain frozen (with respect to backpropagation), and are refined only through Exponential Moving Average (EMA) of the student model's weights {\( \theta_{\text{stud}} \)}. This EMA update gradually integrates target-domain information learned by the student, enabling the teacher to iteratively fine-tune its role in collaborative pseudo-label generation. The EMA update is defined as
\begin{equation} \label{eq:ema}
    \theta_{\text{teach}} \leftarrow \delta \theta_{\text{teach}} + (1 - \delta) \theta_{\text{stud}} ~,
\end{equation}
where \( \delta \) is the decay rate that controls the influence of the student's current weights on the teacher's updated weights.

\vspace*{2mm}
\noindent {\bf{Curriculum Learning.}} 
To effectively balance the contributions of the cross-entropy and discrepancy losses in Eq. \ref{eq:finalLoss}, we employ curriculum learning, which adjusts \( w \) based on the reliability of pseudo-labels. {While most existing approaches rely primarily on domain-specific signals for estimating label reliability \cite{liang2020balanced, qiao2021uncertainty, wang2021uncertainty, CSFDA}, we propose a novel reliability measure that draws upon both domain-specific knowledge and CLIP's domain-agnostic zero-shot capabilities, resulting in a more robust reliability assessment.}

Our approach measures reliability (bounded between 0 and 1) through bidirectional KL divergence between teacher and CLIP predictions, i.e.
\begin{equation}
\label{eq:r}
    r = \exp\left[-\alpha \cdot \frac{\text{KL}(p_{\theta_{\text{teach}}} \parallel p_{\theta_c}) + \text{KL}(p_{\theta_c} \parallel p_{\theta_{\text{teach}}})}{2}\right] ~,
\end{equation}
where \( p_{\theta_{\text{teach}}} \) and \( p_{\theta_c} \) are the predicted probability distributions from the teacher and CLIP models respectively, and \( \alpha \) is a hyperparameter which controls the sensitivity to model disagreement. 
{While the reliability score effectively estimates pseudo-label confidence, optimizing primarily on this metric can lead to suboptimal learning dynamics, potentially overemphasizing high-reliability samples and limiting adaptation to challenging, and yet informative, examples.} 
To address this limitation, we adopt an exponential pace function {(shown in Fig. \ref{Fig:fig-method})}, inspired by \cite{wu2020curricula}, which modulates sample contributions according to both reliability and training progression. Unlike conventional curriculum approaches \cite{peng2021self, wu2020curricula, hacohen2019power}, our method directly applies the pace function to sample weights, not to {ordered data or batch, i.e.}
\begin{equation}
    w = \text{clamp}\left[r \cdot \exp\left(-\beta \cdot e'\right), w_{\text{min}}, w_{\text{max}}\right] ~,   \label{eq-wCL}
\end{equation}
where $e' = e / E$, represents the fraction of completed training (from epoch \( e \) out of a total $E$), \( w_{\text{min}} = 0 \) and \( w_{\text{max}} = 1 \) set bounds for \( w \), and \( \beta \) controls the inclusion rate of lower-reliability samples, with larger values accelerating the participation of these samples in CE loss as training progresses.

\vspace*{2mm}

\noindent {\bf{Adaptive Curriculum Regularization.}}
Despite the effectiveness of curriculum learning in prioritizing reliable pseudo-labels, the model may still suffer from overconfidence, potentially reinforcing early mistakes. To mitigate this, we introduce ACR {to} dynamically adjust sample weights based on prediction patterns, which selectively inverts weights to encourage the model to re-evaluate high-confidence predictions.
{ACR begins with a dynamic weight inversion mechanism, defined as:
\begin{equation}\label{eq:ACR}
    P_{\text{inv}}(e') = 1 - \exp(-\eta \cdot e') ~,
\end{equation}
where \( e' = e/E \) represents the fraction of completed training epochs and \( \eta \in \mathbb{R}^+ \) controls the rate of probability increase. This mechanism ensures a smooth progression of regularization, starting with minimal intervention (\(P_{\text{inv}} \approx 0\) when \(e' = 0\)) and gradually strengthening its effect as training progresses (\(P_{\text{inv}} \approx 1\) when \(e' \rightarrow 1\)). The parameter \( \eta \) determines how aggressively this transition occurs, with larger values leading to earlier and more rapid increases in regularization strength.}
Based on this probability, ACR selects samples for weight adjustment. First, for each sample in a batch of size \( N \), ACR generates uniform random values \( u_i \sim \mathcal{U}(0,1) \) and identifies candidates for inversion by
\begin{equation}\label{eq:pace-exp}
    \mathcal{S} = \left\{ i \in [1, N] : u_i < P_{\text{inv}}(e') \right\}, \quad |\mathcal{S}| \leq \rho N
\end{equation}
where \( \mathcal{S} \) is the set of selected samples for weight inversion, \( P_{\text{inv}}(e') \) represents the current inversion probability, and \( \rho \in [0,1] \) controls the maximum fraction of samples selected in each batch. This process ensures both randomness in sample selection and a controlled upper limit on the number of adjustments per batch.

For each sample in \( \mathcal{S} \), ACR distinguishes between valid high-confidence predictions and those at risk of overconfidence. While high prediction confidence is necessary for effective learning, consistent high confidence combined with subtle fluctuations in the probability distribution might indicate overconfidence. To identify such cases, ACR employs prediction stability analysis and maintains a {history buffer} \( \mathcal{H} \in \mathbb{R}^{X^{T} \times h} \) that stores the last \( h = 10 \) probability distributions \( p_{\theta_{\text{stud}}} \) for each sample in a first-in-first-out manner, and \( X^{T} \) is the number of {target} samples. For each selected sample \( i \), we compute a moving average over its stored distributions to assess stability, with KL divergence between the current probability distribution \( p_{\theta_{\text{stud}}} \) and the moving average \( \bar{p} \) as
\begin{equation}
    D_{\text{KL}}(p_{\theta_{\text{stud}}} \parallel \bar{p}) = \sum_i p_{\theta_{\text{stud}}}(i) \log \frac{p_{\theta_{\text{stud}}}(i)}{\bar{p}(i)}
\end{equation}
{where} $\bar{p} = \frac{1}{h}\sum_{j=1}^{h} p_{\theta_{\text{stud}}}^{(j)}$. 

{The weight adjustment decision for each selected sample then gives the final value for $w$ that is used in Eq. \ref{eq:finalLoss}, i.e,}
\begin{equation}\label{eq:w-update}
    w = \begin{cases} 
        w', & \text{if } D_{\text{KL}} < \sigma \text{ and } c > \gamma \\ 
        w', & \text{if } c < \gamma \\ 
        w, & \text{otherwise (from Eq. \ref{eq-wCL})} 
    \end{cases}
\end{equation}
{where $c = \max(p_{\theta_{\text{stud}}})$ is the current prediction confidence, $ \gamma $ is the confidence threshold (determined based on the maximum confidence of samples where teacher and CLIP predictions agree), and $\sigma$ is a relatively small value $(5\times10^{-2})$ chosen to capture fluctuations in prediction distributions.} In addition to addressing overconfident predictions, weight inversion is also applied to low-confidence samples ($c < \gamma$) to enhance regularization, as model uncertainty may not always indicate incorrect predictions,
\begin{equation}
\label{eq:if}
    w' = (1 - w)\lambda + w(1 - \lambda)
\end{equation}
where \( \lambda \in [0,1] \) is an importance factor that controls the degree of inversion, such that \( \lambda=1 \) applies full inversion, yielding \( w' = 1 - w \), and \( \lambda = 0 \) preserves the original weight as \(w' = w\). This partial inversion mechanism ensures effective regularization while maintaining training stability.

{Details of threshold selection procedures and other hyperparameter settings for curriculum learning and ACR are provided in Supplementary Materials.}
\section{Experiments}


\begin{table*}[h]
\centering \scriptsize
\setlength{\tabcolsep}{2pt}
\renewcommand{\arraystretch}{0.85}
\begin{tabular}{@{}l@{\hspace{0.7em}}l!{\vrule}ccc!{\vrule}ccc!{\vrule}ccc!{\vrule}ccc!{\vrule}c@{}}
\toprule[0.5pt]
& \textbf{Method} & \multicolumn{12}{c}{\textbf{Accuracy (\%)}} & \\
& & \textbf{K→A} & \textbf{K→H} & \textbf{K→M} & \textbf{M→A} & \textbf{M→H} & \textbf{M→K} & \textbf{H→A} & \textbf{H→M} & \textbf{H→K} & \textbf{A→H} & \textbf{A→M} & \textbf{A→K} & \textbf{Avg.} \\
\midrule[0.3pt]

& \cellcolor{blue!15}{DALL-V} \cite{zara2023unreasonable} LB (ResNet50) & \cellcolor{blue!15}15.6 & \cellcolor{blue!15}47.9 & \cellcolor{blue!15}35.7 & \cellcolor{blue!15}34.7 & \cellcolor{blue!15}44.6 & \cellcolor{blue!15}61.6 & \cellcolor{blue!15}17.5 & \cellcolor{blue!15}25.5 & \cellcolor{blue!15}45.1 & \cellcolor{blue!15}14.6 & \cellcolor{blue!15}15.5 & \cellcolor{blue!15}17.8 & \cellcolor{blue!15}31.3 \\
& \cellcolor{blue!15}Co-STAR LB (ResNet50) & \cellcolor{blue!15}26.5 & \cellcolor{blue!15}40.0 & \cellcolor{blue!15}33.5 & \cellcolor{blue!15}35.1 & \cellcolor{blue!15}54.1 & \cellcolor{blue!15}63.6 & \cellcolor{blue!15}22.8 & \cellcolor{blue!15}26.3 & \cellcolor{blue!15}46.2 & \cellcolor{blue!15}15.4 & \cellcolor{blue!15}16.8 & \cellcolor{blue!15}20.7 & \cellcolor{blue!15}33.4 \\
& \cellcolor{blue!15}Co-STAR LB (ViT-B/14) & \cellcolor{blue!15}40.9 & \cellcolor{blue!15}48.8 & \cellcolor{blue!15}46.0 & \cellcolor{blue!15}45.9 & \cellcolor{blue!15}59.2 & \cellcolor{blue!15}74.9 & \cellcolor{blue!15}33.3 & \cellcolor{blue!15}43.5 & \cellcolor{blue!15}54.5 & \cellcolor{blue!15}45.8 & \cellcolor{blue!15}36.3 & \cellcolor{blue!15}44.7 & \cellcolor{blue!15}47.8 \\
\midrule[0.3pt]

\multirow{6}{*}{\rotatebox[origin=c]{90}{\scriptsize\parbox{4em}{\centering\textbf{SFUDA}\\(Image)}}}
& BAIT \cite{BAIT} & 12.7 & 45.7 & 30.0 & 16.9 & 39.6 & 53.0 & 13.6 & 25.5 & 21.2 & 15.7 & 14.5 & 25.5 & 26.2 \\
& MA \cite{MA} [CVPR 2020] & 12.8 & 45.8 & 30.0 & 17.7 & 37.4 & 53.5 & 12.9 & 25.0 & 22.2 & 16.7 & 15.2 & 24.3 & 26.1 \\
& SHOT \cite{SHOT} [ICML 2020] & 12.0 & 44.6 & 29.5 & 15.3 & 36.7 & 51.0 & 13.6 & 24.2 & 21.2 & 17.1 & 14.0 & 24.3 & 25.3 \\
& SHOT++ \cite{SHOT++} & 12.6 & 40.8 & 28.7 & 14.9 & 41.7 & 46.3 & 16.0 & 22.2 & 33.1 & 15.4 & 12.5 & 21.8 & 24.4 \\
& SFDA \cite{SFDA} & 12.6 & 44.9 & 27.5 & 16.0 & 35.2 & 49.2 & 13.1 & 24.2 & 24.9 & 16.3 & 13.2 & 25.2 & 25.2 \\
& CPGA \cite{CPGA} & 13.1 & 46.0 & 30.7 & 18.1 & 39.2 & 55.1 & 13.1 & 26.2 & 25.5 & 19.2 & 16.5 & 26.7 & 26.5 \\
\midrule[0.3pt]

\multirow{6}{*}{\rotatebox[origin=c]{90}{\scriptsize\parbox{4em}{\centering\textbf{SFUVDA}\\(Video)}}}
& ATCoN \cite{ATCON} [ECCV 2022] & 17.2 & 48.2 & 32.5 & \underline{27.2} & 47.3 & 57.7 & 17.9 & 30.7 & 48.5 & 26.7 & 17.2 & 31.0 & 33.5 \\
& STHC \cite{STHC} [CVPR 2023] & 15.5 & 48.7 & 34.8 & 18.4 & 56.3 & 76.6 & 13.8 & 39.8 & 50.1 & 44.6 & 27.3 & 44.7 & 39.2 \\
& DALL-V \cite{zara2023unreasonable} [ICCV 2023] & \underline{24.0} & \underline{52.5} & \underline{47.0} & 24.0 & \textbf{65.4} & \underline{78.1} & 24.0 & \underline{47.0} & \underline{76.7} & \textbf{57.9} & \underline{45.7} & \underline{75.0} & \underline{51.4} \\
& EXTERN \cite{EXTERN} [TMLR 2024] & 23.9 & \textbf{55.8} & 35.2 & 18.1 & 53.7 & 68.1 & \underline{26.2} & 40.7 & 57.6 & 26.2 & 18.2 & 51.4 & 39.6 \\
& Co-STAR (ResNet50) & \textbf{28.2} & \textbf{55.8} & \textbf{47.3} & \textbf{29.0} & \underline{62.5} & \textbf{78.6} & \textbf{28.1} & \textbf{48.0} & \textbf{77.1} & \underline{55.0} & \textbf{46.0} & \textbf{77.8} & \textbf{52.8} \\
& Co-STAR (ViT-B/14) & \textbf{40.2} & \textbf{61.3} & \textbf{60.0} & \textbf{41.7} & \textbf{67.9} & \textbf{85.5} & \textbf{32.7} & \textbf{63.0} & \textbf{87.9} & \textbf{65.0} & \textbf{63.8} & \textbf{88.6} & \textbf{63.1} \\
\midrule[0.3pt]

& \cellcolor{teal!15}{DALL-V} \cite{zara2023unreasonable} UB (ResNet50) & \cellcolor{teal!15}26.9 & \cellcolor{teal!15}70.4 & \cellcolor{teal!15}61.5 & \cellcolor{teal!15}26.9 & \cellcolor{teal!15}70.4 & \cellcolor{teal!15}88.9 & \cellcolor{teal!15}26.9 & \cellcolor{teal!15}61.5 & \cellcolor{teal!15}88.9 & \cellcolor{teal!15}70.4 & \cellcolor{teal!15}61.5 & \cellcolor{teal!15}88.9 & \cellcolor{teal!15}61.9 \\
& \cellcolor{teal!15}Co-STAR UB (ResNet50) & \cellcolor{teal!15}25.8 & \cellcolor{teal!15}66.3 & \cellcolor{teal!15}57.3 & \cellcolor{teal!15}25.8 & \cellcolor{teal!15}66.3 & \cellcolor{teal!15}88.4 & \cellcolor{teal!15}25.8 & \cellcolor{teal!15}57.3 & \cellcolor{teal!15}88.4 & \cellcolor{teal!15}66.3 & \cellcolor{teal!15}57.3 & \cellcolor{teal!15}88.4 & \cellcolor{teal!15}59.4 \\
& \cellcolor{teal!15}Co-STAR UB (ViT-B/14) & \cellcolor{teal!15}58.3 & \cellcolor{teal!15}82.9 & \cellcolor{teal!15}70.3 & \cellcolor{teal!15}58.3 & \cellcolor{teal!15}82.9 & \cellcolor{teal!15}97.5 & \cellcolor{teal!15}58.3 & \cellcolor{teal!15}70.3 & \cellcolor{teal!15}97.5 & \cellcolor{teal!15}82.9 & \cellcolor{teal!15}70.3 & \cellcolor{teal!15}97.5 & \cellcolor{teal!15}77.3 \\
\midrule[0.3pt]

& \multicolumn{14}{c}{{\textbf{Closed Gap (CG, Average \%) across tasks}}} \\
\cmidrule(l{2pt}r{2pt}){2-15}
& DALL-V \cite{zara2023unreasonable} (ResNet50) & -- & 47.5 & 56.7 & -- & \textbf{92.6} & 58.5 & 40.0 & 66.8 & 72.3 & \textbf{83.5} & 71.4 & 80.2 & 67.0 \\
& Co-STAR (ResNet50) & -- & \textbf{60.1} & \textbf{58.0} & -- & 68.9 & \textbf{60.5} & \textbf{100.0} & \textbf{70.0} & \textbf{73.2} & 77.8 & \textbf{72.1} & \textbf{84.3} & \textbf{72.5} \\

\bottomrule[0.5pt]
\end{tabular}
\caption{{{\bf Performance on \textbf{Daily-DA} dataset --} Symbols `H', `K', `A', and `M' represent HMDB51, Kinetics, ARID, and MIT datasets. LB = Lower Bound, UB = Upper Bound. Best results are in \textbf{Bold}, and the second best is \underline{underlined}. {Closed Gap (CG, \%) is also reported, quantifying how much of the LB--UB gap is closed (capped at 100\%). Entries marked with “--” indicate cases where LB exceeds UB, making CG undefined.}}}
\label{tab:Daily-DA}
\end{table*}

   
   

\noindent{\bf{Dataset.}} 
{For our evaluation, we use the same benchmarks as in previous SFUVDA works~\cite{zara2023unreasonable, ATCON, EXTERN}: (i) Daily-DA~\cite{ATCON} comprises 18,949 videos across 8 overlapping action classes from HMDB51~\cite{kuehne2011hmdb}, ARID~\cite{xu2021arid}, MIT~\cite{monfort2019moments}, and Kinetics~\cite{kay2017kinetics}. This benchmark is particularly challenging due to ARID's low-illumination videos, which were specifically filmed under dark conditions, (ii) Sports-DA~\cite{xu2021multi} consists of 40,718 videos spanning 23 overlapping action classes from UCF101~\cite{soomro2012ucf101}, Sports-1M~\cite{karpathy2014large}, and Kinetics~\cite{kay2017kinetics}, primarily focusing on sports activities, and finally (iii) UCF-HMDB\textsubscript{full}~\cite{chen2019temporal} includes 3,209 videos from HMDB51~\cite{kuehne2011hmdb} and UCF101~\cite{soomro2012ucf101} covering 12 action categories.}



\noindent\textbf{Implementation Details.} We follow the same frame sampling strategy for both training and validation as in \cite{zara2023unreasonable}. {We uniformly divide each video into 16 segments and select one random frame per segment during training and the middle frame per segment during validation. Each sampled frame is resized to $224\times224$ pixels before being input to the network.}

For network architecture, we utilize CLIP's visual encoders (either ResNet50 \cite{he2016deep} or {ViT-B/14} \cite{dosovitskiy2021an}) for frame-level feature extraction, followed by a randomly initialized TRN \cite{zhou2018temporal} for both teacher and student models. For training the student model, we use the AdamW optimizer \cite{loshchilov2017decoupled} with a learning rate of 0.001 and a weight decay of 0.2. The teacher model is updated using an 
{EMA with a decay rate of $\delta = 0.999$ (Eq.~\ref{eq:ema}).} The temperature {parameter (Eq.~\ref{eq:kl}) is set empirically at $\tau = 2.0$,  across all datasets.} {Similar to DALL-V \cite{zara2023unreasonable}, we train our model for 30 epochs in all experiments.}
{Additional details, including zero-shot prediction with CLIP, are provided in the Supplementary Materials.}

\noindent{\bf{Comparative Evaluation.}}  
Our comparative results on the Daily-DA, Sports-DA, and HMDB-UCF benchmarks are shown in Tables~\ref{tab:Daily-DA}, \ref{tab:Sports-DA}, and \ref{tab:HMDB}, respectively. Unlike previous works in SFUVDA, which focus solely on ResNet50 backbones, we extend our evaluation to include both ResNet50 and {ViT-B/14} backbones to explore the potential of visual transformers in SFUVDA.

{We report both {\it lower and upper bounds} (LB, UB) for Co-STAR (and DALL-V~\cite{zara2023unreasonable} for context), where LB is the source-only performance on target data and UB is the fully supervised target performance.  
Alongside raw accuracies, we report the \emph{Closed Gap (CG)} metric \cite{yang2021st3d, hu2023density} for fair comparison with methods such as DALL-V that provide LB and UB. CG quantifies how much of the LB--UB gap a method closes, computed as $\text{CG} = \min\left(100,\;\frac{\text{Method} - \text{LB}}{\text{UB} - \text{LB}} \times 100\right)$. A value of 100 means the method reaches or surpasses the UB, while 0 means no improvement over LB.}

{On the Daily-DA benchmark (Table~\ref{tab:Daily-DA}), our method with the ResNet50 backbone achieves the highest average accuracy of 52.8\% across the 12 settings. In particular, when ARID (A) is the target domain, Co-STAR significantly outperforms DALL-V; for example, in the M$\rightarrow$A setting, we obtain 5.0\% higher accuracy. This highlights Co-STAR’s ability to address the substantial domain gap posed by ARID’s low-illumination videos. Using the ViT-B/14 backbone provides a further boost, reaching 63.1\% average accuracy and demonstrating the advantages of transformer architectures for handling complex shifts.}

{With the CG metric, Co-STAR (ResNet50) achieves 72.5\% average versus DALL-V’s 67.0\%, confirming more effective adaptation. Notably, in some settings (e.g., K$\rightarrow$A and M$\rightarrow$A), the lower bound exceeds the upper bound, as rich transferable features from large source domains yield better performance on ARID than fully supervised training on this small, challenging target domain. In these cases, the CG metric becomes undefined or negative. Following prior practice~\cite{hu2023density}, we cap CG within 0--100\%; in cases where the lower bound exceeds the upper bound, CG is undefined and marked as ``--'' to reflect the anomaly.}

On the Sports-DA benchmark (Table~\ref{tab:Sports-DA}), Co-STAR with the ResNet50 backbone {again} sets a new state-of-the-art with an average accuracy of $85.6\%$, surpassing the previous best method, EXTERN, by $2.4\%$. Using the ViT-B/14 backbone further improves performance significantly, achieving average accuracy $92.4\%$. {With CG, Co-STAR obtains 79.2\%, significantly higher than DALL-V's 59.4\%, highlighting significant 
improvement over the LB baseline.
}

{For UCF-HMDB\textsubscript{full} (Table~\ref{tab:HMDB}), Co-STAR (ResNet50)  performs competitively, reaching 90.0\% average accuracy, comparable to leading methods such as STHC (91.5\%) and DALL-V (91.0\%). With CG, Co-STAR achieves 95.7\%, slightly higher than DALL-V’s 95.5\%.
Our ViT-B/14 variant further elevates Co-STAR's performance, achieving an average of 92.6\% and establishing a new state of the art on this benchmark.
}

\begin{table*}[h]
\centering
\scriptsize
\setlength{\tabcolsep}{3.5pt}
\renewcommand{\arraystretch}{0.9}
\begin{tabular}{@{}l@{\hspace{0.8em}}l!{\vrule}cc!{\vrule}cc!{\vrule}cc!{\vrule}c@{}}
\toprule[0.5pt]
& {\bf Method} & \multicolumn{6}{c}{\bf Accuracy (\%)} & \\
& & \textbf{K→U} & \textbf{K→S} & \textbf{S→U} & \textbf{S→K} & \textbf{U→K} & \textbf{U→S} & \textbf{Avg.} \\
\midrule[0.3pt]

& \cellcolor{blue!15}{DALL-V} \cite{zara2023unreasonable} LB (ResNet50) & \cellcolor{blue!15}85.4 & \cellcolor{blue!15}79.5 & \cellcolor{blue!15}84.4 & \cellcolor{blue!15}78.2 & \cellcolor{blue!15}67.2 & \cellcolor{blue!15}64.3 & \cellcolor{blue!15}76.5 \\
& \cellcolor{blue!15}Co-STAR LB (ResNet50) & \cellcolor{blue!15}83.3 & \cellcolor{blue!15}74.3 & \cellcolor{blue!15}83.0 & \cellcolor{blue!15}70.4 & \cellcolor{blue!15}46.7 & \cellcolor{blue!15}44.3 & \cellcolor{blue!15}67.0 \\
& \cellcolor{blue!15}Co-STAR LB (ViT-B/14) & \cellcolor{blue!15}97.7 & \cellcolor{blue!15}88.9 & \cellcolor{blue!15}99.0 & \cellcolor{blue!15}89.0 & \cellcolor{blue!15}86.4 & \cellcolor{blue!15}87.5 & \cellcolor{blue!15}91.4 \\
\midrule[0.3pt]

\multirow{6}{*}{\rotatebox[origin=c]{90}{\scriptsize\parbox{4em}{\centering\textbf{SFUDA}\\(Image)}}}
& BAIT \cite{BAIT} & 92.3 & 66.6 & 88.3 & 72.8 & 57.2 & 44.7 & 70.3 \\
& MA \cite{MA} [CVPR 2020] & 91.0 & 65.9 & 87.8 & 71.9 & 60.7 & 39.4 & 69.5 \\
& SHOT \cite{SHOT} [ICML 2020] & 91.2 & 64.9 & 88.8 & 72.0 & 53.9 & 43.6 & 69.1 \\
& SHOT++ \cite{SHOT++} & 90.0 & 63.1 & 88.0 & 70.3 & 44.7 & 40.9 & 66.2 \\
& SFDA \cite{SFDA} & 86.1 & 60.0 & 85.4 & 68.0 & 55.8 & 43.6 & 66.5 \\
& CPGA \cite{CPGA} & 89.4 & 66.3 & 86.5 & 72.5 & 55.2 & 44.5 & 69.1 \\
\midrule[0.3pt]

\multirow{6}{*}{\rotatebox[origin=c]{90}{\scriptsize\parbox{4em}{\centering\textbf{SFUVDA}\\(Video)}}}
& ATCoN \cite{ATCON} [ECCV 2022] & \underline{93.6} & 69.7 & 90.6 & 76.0 & 65.2 & 47.9 & 73.8 \\
& DALL-V \cite{zara2023unreasonable} [ICCV 2023] & 88.0 & \underline{77.7} & 88.8 & \underline{82.3} & \underline{81.2} & \underline{75.9} & 82.3 \\
& EXTERN \cite{EXTERN} [TMLR 2024] & \textbf{93.7} & 73.8 & \textbf{95.4} & 82.2 & \underline{81.2} & 72.7 & \underline{83.2} \\
& Co-STAR (ResNet50) & 91.2 & \textbf{80.6} & \underline{92.8} & \textbf{84.6} & \textbf{84.5} & \textbf{79.9} & \textbf{85.6} \\
& Co-STAR (ViT-B/14) & \textbf{97.9} & \textbf{90.1} & \textbf{99.3} & \textbf{94.1} & \textbf{90.4} & \textbf{82.4} & \textbf{92.4} \\
\midrule[0.3pt]

& \cellcolor{teal!15}{DALL-V} \cite{zara2023unreasonable} UB (ResNet50) & \cellcolor{teal!15}93.4 & \cellcolor{teal!15}88.3 & \cellcolor{teal!15}93.4 & \cellcolor{teal!15}85.6 & \cellcolor{teal!15}85.6 & \cellcolor{teal!15}88.3 & \cellcolor{teal!15}89.1 \\
& \cellcolor{teal!15}Co-STAR UB (ResNet50) & \cellcolor{teal!15}92.0 & \cellcolor{teal!15}91.5 & \cellcolor{teal!15}92.0 & \cellcolor{teal!15}88.0 & \cellcolor{teal!15}88.0 & \cellcolor{teal!15}91.5 & \cellcolor{teal!15}90.5 \\
& \cellcolor{teal!15}Co-STAR UB (ViT-B/14) & \cellcolor{teal!15}99.4 & \cellcolor{teal!15}97.3 & \cellcolor{teal!15}99.4 & \cellcolor{teal!15}97.4 & \cellcolor{teal!15}97.4 & \cellcolor{teal!15}97.3 & \cellcolor{teal!15}97.9 \\
\midrule[0.3pt]

& \multicolumn{8}{c}{{\textbf{Closed Gap (CG, Average \%) across tasks}}} \\
\cmidrule(l{2pt}r{2pt}){2-9}
& DALL-V \cite{zara2023unreasonable} (ResNet50) & 54.0 & 19.8 & 64.4 & 67.6 & 83.5 & 66.9 & 59.4 \\
& Co-STAR (ResNet50) & \textbf{90.8} & \textbf{36.6} & \textbf{100.0} & \textbf{80.7} & \textbf{91.5} & \textbf{75.4} & \textbf{79.2} \\

\bottomrule[0.5pt]
\end{tabular}
\caption{{{\bf Performance on \textbf{Sports-DA} --} Symbols `K', `U', and `S' represent Kinetics, UCF101, and Sports-1M datasets. LB = Lower Bound, UB = Upper Bound. Best results are in \textbf{Bold}, and the second best is \underline{underlined}. {Closed Gap (CG, \%) is also reported, quantifying how much of the LB--UB performance gap is closed (capped at 100\%).}}}
\label{tab:Sports-DA}
\end{table*}

\begin{table}[h]
\centering
\scriptsize
\setlength{\tabcolsep}{3.5pt}
\renewcommand{\arraystretch}{0.9}
\begin{tabular}{@{}l@{\hspace{0.8em}}l!{\vrule}c!{\vrule}c!{\vrule}c@{}}
\toprule[0.5pt]
& \textbf{Method} & \multicolumn{3}{c}{\textbf{Accuracy (\%)}} \\
& & \textbf{H→U} & \textbf{U→H} & \textbf{Avg.} \\
\midrule[0.3pt]

& \cellcolor{blue!15}{DALL-V} \cite{zara2023unreasonable} LB (ResNet50) & \cellcolor{blue!15}71.6 & \cellcolor{blue!15}76.1 & \cellcolor{blue!15}73.8 \\
& \cellcolor{blue!15}Co-STAR LB (ResNet50) & \cellcolor{blue!15}79.0 & \cellcolor{blue!15}72.8 & \cellcolor{blue!15}75.9 \\
& \cellcolor{blue!15}Co-STAR LB (ViT-B/14) & \cellcolor{blue!15}92.1 & \cellcolor{blue!15}86.8 & \cellcolor{blue!15}89.4 \\
\midrule[0.3pt]

\multirow{6}{*}{\rotatebox[origin=c]{90}{\scriptsize\parbox{4em}{\centering\textbf{SFUDA}\\(Image)}}}
& BAIT \cite{BAIT} & 75.3 & 76.3 & 75.8 \\
& MA \cite{MA} [CVPR 2020] & 74.4 & 67.3 & 70.9 \\
& SHOT \cite{SHOT} [ICML 2020] & 74.4 & 74.4 & 74.4 \\
& SHOT++ \cite{SHOT++} & 71.1 & 68.1 & 69.6 \\
& SFDA \cite{SFDA} & 69.8 & 75.0 & 72.4 \\
& CPGA \cite{CPGA} & 75.8 & 68.1 & 72.0 \\
\midrule[0.3pt]

\multirow{6}{*}{\rotatebox[origin=c]{90}{\scriptsize\parbox{4em}{\centering\textbf{SFUVDA}\\(Video)}}}
& ATCoN \cite{ATCON} [ECCV 2022] & 85.3 & 79.7 & 82.5 \\
& STHC \cite{STHC} [CVPR2023] & \underline{92.1} & \textbf{90.9} & \textbf{91.5} \\
& DALL-V \cite{zara2023unreasonable} [ICCV 2023] & \textbf{93.1} & 88.9 & \underline{91.0} \\
& EXTERN \cite{EXTERN} [TMLR 2024] & 91.9 & 88.9 & 90.4 \\
& Co-STAR (ResNet50) & 90.1 & \underline{89.6} & 90.0 \\
& Co-STAR (ViT-B/14) & \textbf{94.1} & \textbf{92.4} & \textbf{92.6} \\
\midrule[0.3pt]

& \cellcolor{teal!15}{DALL-V} \cite{zara2023unreasonable} UB (ResNet50) & \cellcolor{teal!15}93.7 & \cellcolor{teal!15}91.4 & \cellcolor{teal!15}92.6 \\
& \cellcolor{teal!15}Co-STAR UB (ResNet50) & \cellcolor{teal!15}90.5 & \cellcolor{teal!15}90.5 & \cellcolor{teal!15}90.5 \\
& \cellcolor{teal!15}Co-STAR UB (ViT-B/14) & \cellcolor{teal!15}99.1 & \cellcolor{teal!15}97.5 & \cellcolor{teal!15}98.3 \\
\midrule[0.3pt]

& \multicolumn{4}{c}{{\textbf{Closed Gap (CG, Average \%) across tasks}}} \\
\cmidrule(l{2pt}r{2pt}){2-5}
& DALL-V \cite{zara2023unreasonable} (ResNet50) & \textbf{100.0} & 91.0 & 95.5 \\
& Co-STAR (ResNet50) & 96.5 & \textbf{94.9} & \textbf{95.7} \\
\bottomrule[0.5pt]
\end{tabular}
\caption{{{\bf Performance on UCF-HMDB\textsubscript{full}} -- Symbols `H' and `U' represent HMDB51 and UCF101, respectively. LB = Lower Bound, UB = Upper Bound. Best results are in \textbf{Bold}, and the second best is \underline{underlined}. {Closed Gap (CG, \%) is also reported, quantifying how much of the LB--UB performance gap is closed (capped at 100\%).}}}
\label{tab:HMDB}
\end{table}

\begin{table*}[t]
\centering
\scriptsize
\setlength{\tabcolsep}{4pt}
\renewcommand{\arraystretch}{0.9}
\begin{tabular}{@{}c@{\hspace{0.8em}}cccc|cccc|c|c@{}}
\toprule[0.5pt]
\multicolumn{5}{@{}c}{\bf Components} & \multicolumn{5}{c}{\bf Accuracy (\%)}  \\
Teacher & CLIP & Curr & ACR & & K→Any & M→Any & H→Any & A→Any & Avg. & $\Delta$\\
\midrule
\checkmark & & & & & 45.7 & 61.9 & 42.6 & 54.8 & 51.3 & - \\
 & \checkmark & & & & {\bf 54.0} & {58.3} & {58.3} & {65.1} & {58.9} & - \\

\checkmark & \checkmark & & & & 53.5 & 62.6 & 58.6 &68.7 & 61.0 &+9.7 \\
\checkmark & \checkmark & \checkmark & & & 53.1 & 64.3 & 60.1 & 70.2 & 61.9 & +10.6 \\
\checkmark & \checkmark & \checkmark & \checkmark & & {53.8} & \textbf{65.0} & \textbf{61.2}& \textbf{72.4} & \textbf{63.1} & \textbf{+11.8} \\
\bottomrule[0.5pt]
\end{tabular}
\caption{Ablation study showing the impact of our proposed components on \textbf{Daily-DA}. Results are averaged across target domains for each source domain. $\Delta$ shows absolute improvement over baseline {teacher model}.}
\label{tab:ablation}
\end{table*}

\begin{table}[t]
\scriptsize
\centering
\label{tab:pace_ablation}
\begin{tabular}{lccc}
\toprule
\textbf{Pace Function} & A→H & H→A & \textbf{Avg.} \\
\midrule
Baseline & 60.8 & 31.9 & 46.4 \\
\midrule
Linear & 62.9 & 30.9 & 46.9 \\
Stepwise & 62.9 & 31.9 & 47.4 \\
Sigmoid & 61.3 & 31.9 & 46.6 \\
Exponential & \textbf{65.0} & \textbf{32.7} & \textbf{48.9} \\
\bottomrule
\end{tabular}
\caption{Ablation study of different pace functions in curriculum learning. Best accuracy (\%) is reported for each dataset. The baseline represents a setting without any pace function, where only the reliability score from Eq.~\ref{eq:r} is used to balance between losses.}
\label{tab:pace}
\end{table}

\section{Ablation Study and Analysis}
\label{ablation_studies}
To analyze the effectiveness of each component in our proposed Co-STAR framework, we conduct comprehensive ablation experiments on the Daily-DA dataset using ViT-B/14 as the backbone network. Table~\ref{tab:ablation} summarises the results, where we gradually incorporate different components and evaluate their impact on adaptation performance.

\noindent\textbf{Baseline Model.} We first evaluate the performance using only the teacher model for self-training (first row of Table~\ref{tab:ablation}), which achieves an average accuracy of $51.3\%$ across all domain adaptations. While this serves as a reasonable baseline, it highlights the limitations of relying solely on a source-trained model for adaptation.

\noindent\textbf{Impact of CLIP Integration.} As shown in Table~\ref{tab:ablation} {(row 3)}, adding CLIP for collaborative pseudo-labeling significantly improves performance, increasing average accuracy by $9.7\%$ to $61.0\%$. {While CLIP alone achieves respectable zero-shot performance ($58.9\%$, row 2), the collaborative approach outperforms both individual models.}  This substantial gain demonstrates the effectiveness of leveraging CLIP's domain-agnostic knowledge alongside the teacher's domain-specific features. The improvement is particularly pronounced for challenging adaptations, with H$\rightarrow$Any seeing a $16.0$ point increase from $42.6\%$ to $58.6\%$) and A$\rightarrow$Any improving by $13.9$ points from $54.8\%$ to  $68.7\%$.

\noindent\textbf{Impact of Curriculum Learning.} Incorporating our curriculum learning strategy further enhances performance to $61.9\%$ ($+10.6\%$ over baseline), as evidenced in Table~\ref{tab:ablation}. This improvement validates our approach of dynamically weighting samples based on prediction reliability.

\noindent\textbf{Impact of ACR.} Finally, Table~\ref{tab:ablation} shows that adding ACR to create the complete Co-STAR framework yields the best performance with 63.1\% average accuracy, representing an 11.8 percentage point improvement over the baseline. The strongest gains are observed in A$\rightarrow$Any adaptations which improves from 70.2\% to 72.4\%.

\begin{figure}[h]
\centerline{\includegraphics[scale=0.145]{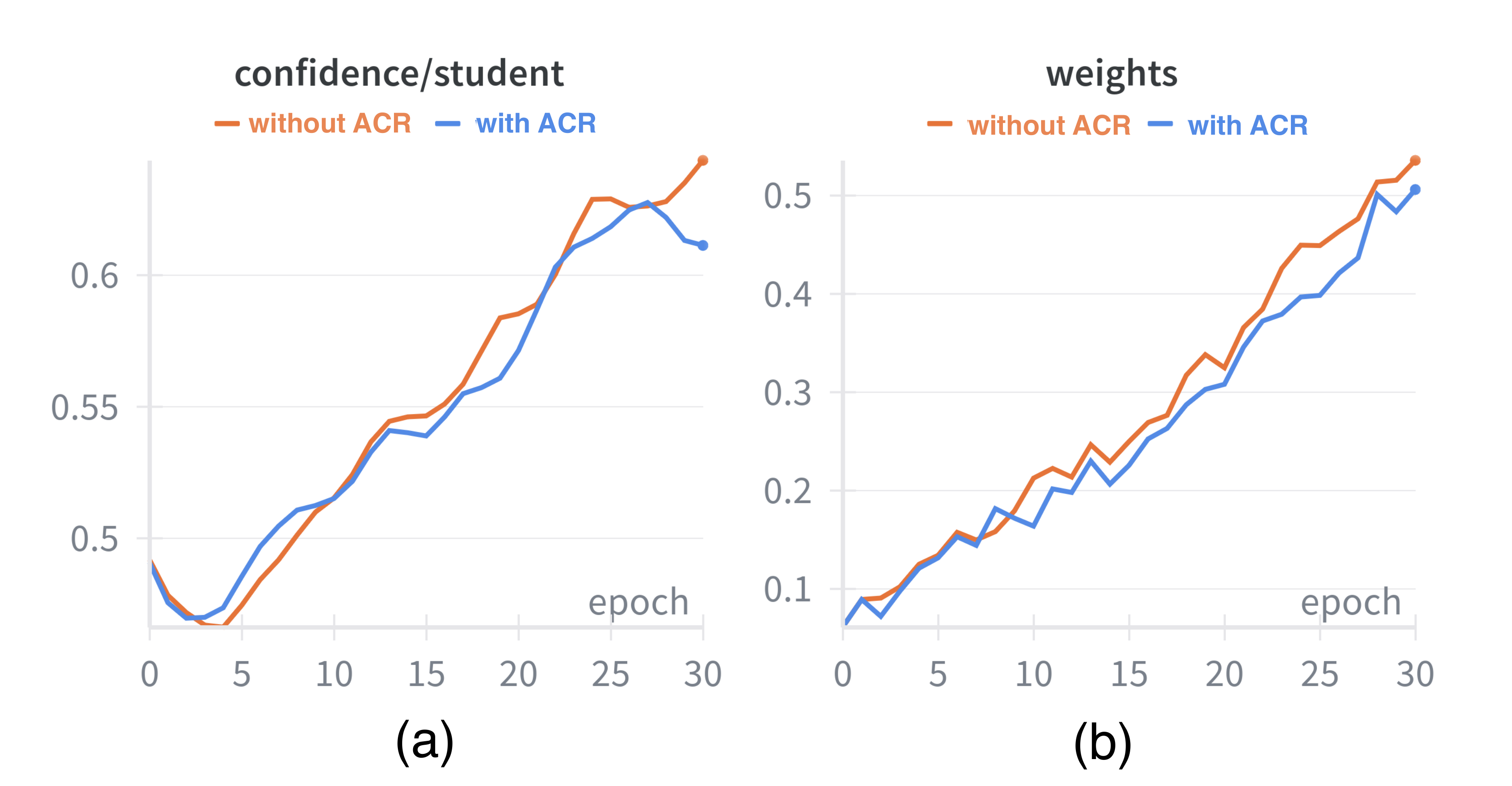}}
\caption{Training dynamics with and without ACR for A$\rightarrow$H adaptation. (a) Student confidence over epochs, computed as epoch-averaged maximum softmax probability per prediction. (b) Curriculum weights over epochs showing the epoch-averaged minimum weight across all training batches.}

\label{Fig:training_dyn} 
\end{figure}

{To further illustrate ACR's effectiveness, Fig.~\ref{Fig:training_dyn} presents the training dynamics for A$\rightarrow$H adaptation. Fig.~\ref{Fig:training_dyn}(a) shows student confidence evolution, calculated as epoch-averaged maximum softmax probability across classes for each prediction. Without ACR, student confidence grows unchecked and reaches peak levels around epoch 25 and continuing to rise which indicates overconfident predictions. In contrast, ACR actively intervenes by allowing confidence to build gradually but then reducing it in later epochs when potential overconfidence is detected. Fig.~\ref{Fig:training_dyn}(b) tracks curriculum weights across training, where we plot the epoch-averaged minimum weight across all batches to highlight how the model treats its most challenging samples. Both methods follow similar exponential growth patterns as expected from our curriculum design, but ACR maintains slightly lower values for the hardest samples. This indicates that ACR keeps unreliable samples at lower influence levels while still allowing proper curriculum progression from knowledge distillation to pseudo-label reliance. Together, these dynamics validate that ACR successfully addresses overconfidence without disrupting the fundamental curriculum learning mechanism.}


\noindent\textbf{Effectiveness of Pace Functions.} {Table~\ref{tab:pace} compares four different pace functions and a baseline (i.e. no pace function, using only reliability scores from Eq.~\ref{eq:r}) on adaptation scenarios between HMDB51 and Arid datasets in Daily-DA benchmark.} The exponential pace function achieves the best performance with 48.85\% average accuracy (65.0\% on A→H and 32.7\% on H→A), outperforming linear (46.9\%), stepwise (47.4\%), and sigmoid (46.6\%) functions. The superior performance of exponential pacing can be attributed to two key characteristics: its gradual increase in early stages ensures the model primarily relies on the most reliable pseudo-labels when the adaptation process is most sensitive to noise, unlike linear and stepwise functions that might introduce less reliable samples too quickly. In addition, the accelerated inclusion of samples in later stages, when the model has developed more robust features, allows for more aggressive learning. This is in contrast to sigmoid's symmetric structure, which levels off in later stages and may under-utilize the model's improved capacity for learning from less reliable samples. Detailed formulations of all pace functions are in Supplementary Materials.

\section{Limitations and Future Work} 
The promising results of Co-STAR, are somewhat dependent on the zero-shot capability of CLIP, which may reduce effectiveness for action classes or scenarios that are under-represented in CLIP's training data. This reliance can limit the consistency of our collaborative pseudo-labeling strategy in such cases. Our evaluation also focuses on action recognition alone, but the method could be evaluated on other video understanding tasks with large domain gaps, such as action quality assessment \cite{dadashzadeh2024pecop}, temporal action localization \cite{xia2020survey}, {and video object segmentation }\cite{gao2023deep}. These tasks involve subtle cross-domain differences and could further reveal Co-STAR's generalizability. Future work will aim to address these limitations.

\section{Conclusions}
{In this work, we propose Co-STAR, a novel source-free unsupervised video domain adaptation framework that addresses the challenges of noisy pseudo-labels and overconfident predictions through an integrated approach of collaborative self-training and curriculum learning. Co-STAR combines a source-trained teacher model with the zero-shot capabilities of CLIP, using a reliability-based weighting mechanism to balance confident and uncertain predictions. 
By incorporating adaptive curriculum regularization, our framework dynamically adjusts sample weights based on prediction stability, effectively reducing overfitting and enhancing domain adaptation. Our extensive experiments using both CNN-based and visual transformer-based backbones demonstrate that Co-STAR achieves state-of-the-art performance across three video domain adaptation benchmarks.}

\section{Acknowledgements}
This work was supported by the TORUS Project, which has been funded by the UK Engineering and Physical Sciences Research Council (EPSRC), grant number EP/X036146/1.

{
    \small
    \bibliographystyle{ieeenat_fullname}
    \bibliography{main}

\begin{thebibliography}{56}
\providecommand{\natexlab}[1]{#1}
\providecommand{\url}[1]{\texttt{#1}}
\expandafter\ifx\csname urlstyle\endcsname\relax
  \providecommand{\doi}[1]{doi: #1}\else
  \providecommand{\doi}{doi: \begingroup \urlstyle{rm}\Url}\fi

\bibitem[Chen et~al.(2019)Chen, Kira, AlRegib, Yoo, Chen, and Zheng]{chen2019temporal}
Min-Hung Chen, Zsolt Kira, Ghassan AlRegib, Jaekwon Yoo, Ruxin Chen, and Jian Zheng.
\newblock Temporal attentive alignment for large-scale video domain adaptation.
\newblock In \emph{Proceedings of the IEEE/CVF International Conference on Computer Vision}, pages 6321--6330, 2019.

\bibitem[Chen et~al.(2021)Chen, Li, Wu, Dong, and Shi]{chen2021unsupervised}
Pengfei Chen, Leida Li, Jinjian Wu, Weisheng Dong, and Guangming Shi.
\newblock Unsupervised curriculum domain adaptation for no-reference video quality assessment.
\newblock In \emph{Proceedings of the IEEE/CVF International Conference on Computer Vision}, pages 5178--5187, 2021.

\bibitem[Choi et~al.(2019)Choi, Jeong, Kim, and Kim]{choi2019pseudo}
Jaehoon Choi, Minki Jeong, Taekyung Kim, and Changick Kim.
\newblock Pseudo-labeling curriculum for unsupervised domain adaptation.
\newblock \emph{arXiv preprint arXiv:1908.00262}, 2019.

\bibitem[Da~Costa et~al.(2022{\natexlab{a}})Da~Costa, Zara, Rota, Oliveira-Santos, Sebe, Murino, and Ricci]{da2022dual}
Victor G~Turrisi Da~Costa, Giacomo Zara, Paolo Rota, Thiago Oliveira-Santos, Nicu Sebe, Vittorio Murino, and Elisa Ricci.
\newblock Dual-head contrastive domain adaptation for video action recognition.
\newblock In \emph{Proceedings of the IEEE/CVF Winter Conference on Applications of Computer Vision}, pages 1181--1190, 2022{\natexlab{a}}.

\bibitem[Da~Costa et~al.(2022{\natexlab{b}})Da~Costa, Zara, Rota, Oliveira-Santos, Sebe, Murino, and Ricci]{da2022unsupervised}
Victor G~Turrisi Da~Costa, Giacomo Zara, Paolo Rota, Thiago Oliveira-Santos, Nicu Sebe, Vittorio Murino, and Elisa Ricci.
\newblock Unsupervised domain adaptation for video transformers in action recognition.
\newblock In \emph{2022 26th International Conference on Pattern Recognition (ICPR)}, pages 1258--1265. IEEE, 2022{\natexlab{b}}.

\bibitem[Dadashzadeh et~al.(2024)Dadashzadeh, Duan, Whone, and Mirmehdi]{dadashzadeh2024pecop}
Amirhossein Dadashzadeh, Shuchao Duan, Alan Whone, and Majid Mirmehdi.
\newblock Pecop: Parameter efficient continual pretraining for action quality assessment.
\newblock In \emph{Proceedings of the IEEE/CVF Winter Conference on Applications of Computer Vision}, pages 42--52, 2024.

\bibitem[Dasgupta et~al.(2022)Dasgupta, Jawahar, and Alahari]{dasgupta2022overcoming}
Avijit Dasgupta, CV Jawahar, and Karteek Alahari.
\newblock Overcoming label noise for source-free unsupervised video domain adaptation.
\newblock In \emph{Proceedings of the Thirteenth Indian Conference on Computer Vision, Graphics and Image Processing}, pages 1--9, 2022.

\bibitem[Dosovitskiy et~al.(2021)Dosovitskiy, Beyer, Kolesnikov, Weissenborn, Zhai, Unterthiner, Dehghani, Minderer, Heigold, Gelly, Uszkoreit, and Houlsby]{dosovitskiy2021an}
Alexey Dosovitskiy, Lucas Beyer, Alexander Kolesnikov, Dirk Weissenborn, Xiaohua Zhai, Thomas Unterthiner, Mostafa Dehghani, Matthias Minderer, Georg Heigold, Sylvain Gelly, Jakob Uszkoreit, and Neil Houlsby.
\newblock An image is worth 16x16 words: Transformers for image recognition at scale.
\newblock In \emph{International Conference on Learning Representations}, 2021.

\bibitem[Fang et~al.(2024)Fang, Yap, Lin, Zhu, and Liu]{fang2024source}
Yuqi Fang, Pew-Thian Yap, Weili Lin, Hongtu Zhu, and Mingxia Liu.
\newblock Source-free unsupervised domain adaptation: A survey.
\newblock \emph{Neural Networks}, page 106230, 2024.

\bibitem[Feichtenhofer et~al.(2016)Feichtenhofer, Pinz, and Zisserman]{feichtenhofer2016convolutional}
Christoph Feichtenhofer, Axel Pinz, and Andrew Zisserman.
\newblock Convolutional two-stream network fusion for video action recognition.
\newblock In \emph{Proceedings of the IEEE conference on computer vision and pattern recognition}, pages 1933--1941, 2016.

\bibitem[Feichtenhofer et~al.(2017)Feichtenhofer, Pinz, and Wildes]{feichtenhofer2017spatiotemporal}
Christoph Feichtenhofer, Axel Pinz, and Richard~P Wildes.
\newblock Spatiotemporal multiplier networks for video action recognition.
\newblock In \emph{Proceedings of the IEEE conference on computer vision and pattern recognition}, pages 4768--4777, 2017.

\bibitem[Gao et~al.(2023)Gao, Zheng, Yu, Shan, Ding, and Han]{gao2023deep}
Mingqi Gao, Feng Zheng, James~JQ Yu, Caifeng Shan, Guiguang Ding, and Jungong Han.
\newblock Deep learning for video object segmentation: a review.
\newblock \emph{Artificial Intelligence Review}, 56\penalty0 (1):\penalty0 457--531, 2023.

\bibitem[Hacohen and Weinshall(2019)]{hacohen2019power}
Guy Hacohen and Daphna Weinshall.
\newblock On the power of curriculum learning in training deep networks.
\newblock In \emph{International conference on machine learning}, pages 2535--2544. PMLR, 2019.

\bibitem[Han et~al.(2022)Han, Wang, Chen, Chen, Guo, Liu, Tang, Xiao, Xu, Xu, et~al.]{han2022survey}
Kai Han, Yunhe Wang, Hanting Chen, Xinghao Chen, Jianyuan Guo, Zhenhua Liu, Yehui Tang, An Xiao, Chunjing Xu, Yixing Xu, et~al.
\newblock A survey on vision transformer.
\newblock \emph{IEEE transactions on pattern analysis and machine intelligence}, 45\penalty0 (1):\penalty0 87--110, 2022.

\bibitem[He et~al.(2016)He, Zhang, Ren, and Sun]{he2016deep}
Kaiming He, Xiangyu Zhang, Shaoqing Ren, and Jian Sun.
\newblock Deep residual learning for image recognition.
\newblock In \emph{Proceedings of the IEEE conference on computer vision and pattern recognition}, pages 770--778, 2016.

\bibitem[Htet et~al.(2022)Htet, Zin, Tin, Tamura, Kondo, and Chosa]{htet2022hmm}
Ye Htet, Thi~Thi Zin, Pyke Tin, Hiroki Tamura, Kazuhiro Kondo, and Etsuo Chosa.
\newblock Hmm-based action recognition system for elderly healthcare by colorizing depth map.
\newblock \emph{International Journal of Environmental Research and Public Health}, 19\penalty0 (19):\penalty0 12055, 2022.

\bibitem[Ishikawa et~al.(2024)Ishikawa, Kondo, and Kataoka]{ishikawa2024learnable}
Yuchi Ishikawa, Masayoshi Kondo, and Hirokatsu Kataoka.
\newblock Learnable cube-based video encryption for privacy-preserving action recognition.
\newblock In \emph{Proceedings of the IEEE/CVF Winter Conference on Applications of Computer Vision}, pages 7003--7013, 2024.

\bibitem[Karim et~al.(2023)Karim, Mithun, Rajvanshi, Chiu, Samarasekera, and Rahnavard]{CSFDA}
Nazmul Karim, Niluthpol~Chowdhury Mithun, Abhinav Rajvanshi, Han-pang Chiu, Supun Samarasekera, and Nazanin Rahnavard.
\newblock C-sfda: A curriculum learning aided self-training framework for efficient source free domain adaptation.
\newblock In \emph{Proceedings of the IEEE/CVF Conference on Computer Vision and Pattern Recognition}, pages 24120--24131, 2023.

\bibitem[Karpathy et~al.(2014)Karpathy, Toderici, Shetty, Leung, Sukthankar, and Fei-Fei]{karpathy2014large}
Andrej Karpathy, George Toderici, Sanketh Shetty, Thomas Leung, Rahul Sukthankar, and Li Fei-Fei.
\newblock Large-scale video classification with convolutional neural networks.
\newblock In \emph{Proceedings of the IEEE conference on Computer Vision and Pattern Recognition}, pages 1725--1732, 2014.

\bibitem[Kay et~al.(2017)Kay, Carreira, Simonyan, Zhang, Hillier, Vijayanarasimhan, Viola, Green, Back, Natsev, et~al.]{kay2017kinetics}
Will Kay, Joao Carreira, Karen Simonyan, Brian Zhang, Chloe Hillier, Sudheendra Vijayanarasimhan, Fabio Viola, Tim Green, Trevor Back, Paul Natsev, et~al.
\newblock The kinetics human action video dataset.
\newblock \emph{arXiv preprint arXiv:1705.06950}, 2017.

\bibitem[Kim et~al.(2021)Kim, Cho, Han, Panda, and Hong]{SFDA}
Youngeun Kim, Donghyeon Cho, Kyeongtak Han, Priyadarshini Panda, and Sungeun Hong.
\newblock Domain adaptation without source data.
\newblock \emph{IEEE Transactions on Artificial Intelligence}, 2\penalty0 (6):\penalty0 508--518, 2021.

\bibitem[Kuehne et~al.(2011)Kuehne, Jhuang, Garrote, Poggio, and Serre]{kuehne2011hmdb}
Hildegard Kuehne, Hueihan Jhuang, Est{\'\i}baliz Garrote, Tomaso Poggio, and Thomas Serre.
\newblock Hmdb: a large video database for human motion recognition.
\newblock In \emph{2011 International conference on computer vision}, pages 2556--2563. IEEE, 2011.

\bibitem[Li et~al.(2023)Li, Patel, Kruus, and Min]{STHC}
Kai Li, Deep Patel, Erik Kruus, and Martin~Renqiang Min.
\newblock Source-free video domain adaptation with spatial-temporal-historical consistency learning.
\newblock In \emph{Proceedings of the IEEE/CVF Conference on Computer Vision and Pattern Recognition}, pages 14643--14652, 2023.

\bibitem[Li et~al.(2020)Li, Jiao, Cao, Wong, and Wu]{MA}
Rui Li, Qianfen Jiao, Wenming Cao, Hau-San Wong, and Si Wu.
\newblock Model adaptation: Unsupervised domain adaptation without source data.
\newblock In \emph{Proceedings of the IEEE/CVF conference on computer vision and pattern recognition}, pages 9641--9650, 2020.

\bibitem[Lian et~al.(2019)Lian, Lv, Duan, and Gong]{lian2019constructing}
Qing Lian, Fengmao Lv, Lixin Duan, and Boqing Gong.
\newblock Constructing self-motivated pyramid curriculums for cross-domain semantic segmentation: A non-adversarial approach.
\newblock In \emph{Proceedings of the IEEE/CVF International Conference on Computer Vision}, pages 6758--6767, 2019.

\bibitem[Liang et~al.(2020{\natexlab{a}})Liang, Hu, and Feng]{SHOT}
Jian Liang, Dapeng Hu, and Jiashi Feng.
\newblock Do we really need to access the source data? source hypothesis transfer for unsupervised domain adaptation.
\newblock In \emph{International conference on machine learning}, pages 6028--6039. PMLR, 2020{\natexlab{a}}.

\bibitem[Liang et~al.(2020{\natexlab{b}})Liang, Wang, Hu, He, and Feng]{liang2020balanced}
Jian Liang, Yunbo Wang, Dapeng Hu, Ran He, and Jiashi Feng.
\newblock A balanced and uncertainty-aware approach for partial domain adaptation.
\newblock In \emph{European conference on computer vision}, pages 123--140. Springer, 2020{\natexlab{b}}.

\bibitem[Liang et~al.(2021)Liang, Hu, Wang, He, and Feng]{SHOT++}
Jian Liang, Dapeng Hu, Yunbo Wang, Ran He, and Jiashi Feng.
\newblock Source data-absent unsupervised domain adaptation through hypothesis transfer and labeling transfer.
\newblock \emph{IEEE Transactions on Pattern Analysis and Machine Intelligence}, 44\penalty0 (11):\penalty0 8602--8617, 2021.

\bibitem[Loshchilov(2017)]{loshchilov2017decoupled}
I Loshchilov.
\newblock Decoupled weight decay regularization.
\newblock \emph{arXiv preprint arXiv:1711.05101}, 2017.

\bibitem[Monfort et~al.(2019)Monfort, Andonian, Zhou, Ramakrishnan, Bargal, Yan, Brown, Fan, Gutfreund, Vondrick, et~al.]{monfort2019moments}
Mathew Monfort, Alex Andonian, Bolei Zhou, Kandan Ramakrishnan, Sarah~Adel Bargal, Tom Yan, Lisa Brown, Quanfu Fan, Dan Gutfreund, Carl Vondrick, et~al.
\newblock Moments in time dataset: one million videos for event understanding.
\newblock \emph{IEEE transactions on pattern analysis and machine intelligence}, 42\penalty0 (2):\penalty0 502--508, 2019.

\bibitem[Paul and Singh(2014)]{paul2014survey}
S~Nissi Paul and Y~Jayanta Singh.
\newblock Survey on video analysis of human walking motion.
\newblock \emph{International Journal of Signal Processing, Image Processing and Pattern Recognition}, 7\penalty0 (3):\penalty0 99--122, 2014.

\bibitem[Peng et~al.(2021)Peng, Wang, Desrosiers, and Pedersoli]{peng2021self}
Jizong Peng, Ping Wang, Christian Desrosiers, and Marco Pedersoli.
\newblock Self-paced contrastive learning for semi-supervised medical image segmentation with meta-labels.
\newblock \emph{Advances in Neural Information Processing Systems}, 34:\penalty0 16686--16699, 2021.

\bibitem[Presti and La~Cascia(2016)]{presti20163d}
Liliana~Lo Presti and Marco La~Cascia.
\newblock 3d skeleton-based human action classification: A survey.
\newblock \emph{Pattern Recognition}, 53:\penalty0 130--147, 2016.

\bibitem[Qiao and Peng(2021)]{qiao2021uncertainty}
Fengchun Qiao and Xi Peng.
\newblock Uncertainty-guided model generalization to unseen domains.
\newblock In \emph{Proceedings of the IEEE/CVF conference on computer vision and pattern recognition}, pages 6790--6800, 2021.

\bibitem[Qiu et~al.(2021)Qiu, Zhang, Lin, Niu, Liu, Du, and Tan]{CPGA}
Zhen Qiu, Yifan Zhang, Hongbin Lin, Shuaicheng Niu, Yanxia Liu, Qing Du, and Mingkui Tan.
\newblock Source-free domain adaptation via avatar prototype generation and adaptation.
\newblock \emph{arXiv preprint arXiv:2106.15326}, 2021.

\bibitem[Radford et~al.(2021)Radford, Kim, Hallacy, Ramesh, Goh, Agarwal, Sastry, Askell, Mishkin, Clark, et~al.]{radford2021learning}
Alec Radford, Jong~Wook Kim, Chris Hallacy, Aditya Ramesh, Gabriel Goh, Sandhini Agarwal, Girish Sastry, Amanda Askell, Pamela Mishkin, Jack Clark, et~al.
\newblock Learning transferable visual models from natural language supervision.
\newblock In \emph{International conference on machine learning}, pages 8748--8763. PMLR, 2021.

\bibitem[Reddy et~al.(2024)Reddy, Paul, Rivera, Shah, de~Melo, and Chellappa]{reddy2024unsupervised}
Arun Reddy, William Paul, Corban Rivera, Ketul Shah, Celso~M de Melo, and Rama Chellappa.
\newblock Unsupervised video domain adaptation with masked pre-training and collaborative self-training.
\newblock In \emph{Proceedings of the IEEE/CVF Conference on Computer Vision and Pattern Recognition}, pages 18919--18929, 2024.

\bibitem[Roy et~al.(2021)Roy, Krivosheev, Zhong, Sebe, and Ricci]{roy2021curriculum}
Subhankar Roy, Evgeny Krivosheev, Zhun Zhong, Nicu Sebe, and Elisa Ricci.
\newblock Curriculum graph co-teaching for multi-target domain adaptation.
\newblock In \emph{Proceedings of the IEEE/CVF conference on computer vision and pattern recognition}, pages 5351--5360, 2021.

\bibitem[Sahoo et~al.(2021)Sahoo, Shah, Panda, Saenko, and Das]{sahoo2021contrast}
Aadarsh Sahoo, Rutav Shah, Rameswar Panda, Kate Saenko, and Abir Das.
\newblock Contrast and mix: Temporal contrastive video domain adaptation with background mixing.
\newblock \emph{Advances in Neural Information Processing Systems}, 34:\penalty0 23386--23400, 2021.

\bibitem[Sakaridis et~al.(2019)Sakaridis, Dai, and Gool]{sakaridis2019guided}
Christos Sakaridis, Dengxin Dai, and Luc~Van Gool.
\newblock Guided curriculum model adaptation and uncertainty-aware evaluation for semantic nighttime image segmentation.
\newblock In \emph{Proceedings of the IEEE/CVF international conference on computer vision}, pages 7374--7383, 2019.

\bibitem[Soomro(2012)]{soomro2012ucf101}
K Soomro.
\newblock Ucf101: A dataset of 101 human actions classes from videos in the wild.
\newblock \emph{arXiv preprint arXiv:1212.0402}, 2012.

\bibitem[Wang et~al.(2021{\natexlab{a}})Wang, Xing, and Liu]{wang2021actionclip}
Mengmeng Wang, Jiazheng Xing, and Yong Liu.
\newblock Actionclip: A new paradigm for video action recognition.
\newblock \emph{arXiv preprint arXiv:2109.08472}, 2021{\natexlab{a}}.

\bibitem[Wang et~al.(2021{\natexlab{b}})Wang, Peng, and Zhang]{wang2021uncertainty}
Yuxi Wang, Junran Peng, and ZhaoXiang Zhang.
\newblock Uncertainty-aware pseudo label refinery for domain adaptive semantic segmentation.
\newblock In \emph{Proceedings of the IEEE/CVF international conference on computer vision}, pages 9092--9101, 2021{\natexlab{b}}.

\bibitem[Wu et~al.(2023)Wu, Sun, and Ouyang]{wu2023revisiting}
Wenhao Wu, Zhun Sun, and Wanli Ouyang.
\newblock Revisiting classifier: Transferring vision-language models for video recognition.
\newblock In \emph{Proceedings of the AAAI conference on artificial intelligence}, pages 2847--2855, 2023.

\bibitem[Wu et~al.(2020)Wu, Dyer, and Neyshabur]{wu2020curricula}
Xiaoxia Wu, Ethan Dyer, and Behnam Neyshabur.
\newblock When do curricula work?
\newblock \emph{arXiv preprint arXiv:2012.03107}, 2020.

\bibitem[Xia and Zhan(2020)]{xia2020survey}
Huifen Xia and Yongzhao Zhan.
\newblock A survey on temporal action localization.
\newblock \emph{IEEE Access}, 8:\penalty0 70477--70487, 2020.

\bibitem[Xu et~al.(2021{\natexlab{a}})Xu, Ghosh, Huang, Okhonko, Aghajanyan, Metze, Zettlemoyer, and Feichtenhofer]{xu2021videoclip}
Hu Xu, Gargi Ghosh, Po-Yao Huang, Dmytro Okhonko, Armen Aghajanyan, Florian Metze, Luke Zettlemoyer, and Christoph Feichtenhofer.
\newblock Videoclip: Contrastive pre-training for zero-shot video-text understanding.
\newblock \emph{arXiv preprint arXiv:2109.14084}, 2021{\natexlab{a}}.

\bibitem[Xu et~al.(2021{\natexlab{b}})Xu, Yang, Cao, Mao, Yin, and See]{xu2021arid}
Yuecong Xu, Jianfei Yang, Haozhi Cao, Kezhi Mao, Jianxiong Yin, and Simon See.
\newblock Arid: A new dataset for recognizing action in the dark.
\newblock In \emph{Deep Learning for Human Activity Recognition: Second International Workshop, DL-HAR 2020, Held in Conjunction with IJCAI-PRICAI 2020, Kyoto, Japan, January 8, 2021, Proceedings 2}, pages 70--84. Springer, 2021{\natexlab{b}}.

\bibitem[Xu et~al.(2021{\natexlab{c}})Xu, Yang, Cao, Wu, Wu, Zhao, and Chen]{xu2021multi}
Yuecong Xu, Jianfei Yang, Haozhi Cao, Keyu Wu, Min Wu, Rui Zhao, and Zhenghua Chen.
\newblock Multi-source video domain adaptation with temporal attentive moment alignment.
\newblock \emph{arXiv preprint arXiv:2109.09964}, 2021{\natexlab{c}}.

\bibitem[Xu et~al.(2022)Xu, Yang, Cao, Wu, Wu, and Chen]{ATCON}
Yuecong Xu, Jianfei Yang, Haozhi Cao, Keyu Wu, Min Wu, and Zhenghua Chen.
\newblock Source-free video domain adaptation by learning temporal consistency for action recognition.
\newblock In \emph{European Conference on Computer Vision}, pages 147--164. Springer, 2022.

\bibitem[Xu et~al.(2024)Xu, Yang, Cao, Wu, Li, Xie, and Chen]{EXTERN}
Yuecong Xu, Jianfei Yang, Haozhi Cao, Min Wu, Xiaoli Li, Lihua Xie, and Zhenghua Chen.
\newblock Leveraging endo- and exo-temporal regularization for black-box video domain adaptation.
\newblock \emph{Transactions on Machine Learning Research}, 2024.

\bibitem[Yang et~al.(2020)Yang, Wang, Van De~Weijer, Herranz, and Jui]{BAIT}
Shiqi Yang, Yaxing Wang, Joost Van De~Weijer, Luis Herranz, and Shangling Jui.
\newblock Unsupervised domain adaptation without source data by casting a bait.
\newblock \emph{arXiv preprint arXiv:2010.12427}, 1\penalty0 (2):\penalty0 5, 2020.

\bibitem[Zara et~al.(2023)Zara, Conti, Roy, Lathuili{\`e}re, Rota, and Ricci]{zara2023unreasonable}
Giacomo Zara, Alessandro Conti, Subhankar Roy, St{\'e}phane Lathuili{\`e}re, Paolo Rota, and Elisa Ricci.
\newblock The unreasonable effectiveness of large language-vision models for source-free video domain adaptation.
\newblock In \emph{Proceedings of the IEEE/CVF International Conference on Computer Vision}, pages 10307--10317, 2023.

\bibitem[Zhang et~al.(2023)Zhang, Shen, and Foo]{zhang2023rethinking}
Wenyu Zhang, Li Shen, and Chuan-Sheng Foo.
\newblock Rethinking the role of pre-trained networks in source-free domain adaptation.
\newblock In \emph{Proceedings of the IEEE/CVF International Conference on Computer Vision}, pages 18841--18851, 2023.

\bibitem[Zhang et~al.(2019)Zhang, David, Foroosh, and Gong]{zhang2019curriculum}
Yang Zhang, Philip David, Hassan Foroosh, and Boqing Gong.
\newblock A curriculum domain adaptation approach to the semantic segmentation of urban scenes.
\newblock \emph{IEEE transactions on pattern analysis and machine intelligence}, 42\penalty0 (8):\penalty0 1823--1841, 2019.

\bibitem[Zhou et~al.(2018)Zhou, Andonian, Oliva, and Torralba]{zhou2018temporal}
Bolei Zhou, Alex Andonian, Aude Oliva, and Antonio Torralba.
\newblock Temporal relational reasoning in videos.
\newblock In \emph{Proceedings of the European conference on computer vision (ECCV)}, pages 803--818, 2018.

\end{thebibliography}
}
\clearpage
\appendix
\section*{\Large{Appendix}}


\begin{algorithm}[b]
\small
\footnotesize
\caption{Zero-Shot video classification using CLIP}
\begin{algorithmic}[1]
\REQUIRE 
    \text{Target domain} $\mathcal{D}_t = \{x^t_i\}_{i=1}^{X_t}$, \text{Temperature parameter} $\tau_c$ = 0.5, \text{CLIP visual encoder} $G_v$, \text{CLIP text encoder} $G_t$
\FOR{$x^t_i$ in $\mathcal{D}_t$}
    \STATE // Extract visual features
    \STATE $v \gets G_v(x^t_i)$
    \STATE $v \gets \texttt{normalize}(v)$
    \STATE $v \gets \texttt{avg\_p}(v)$ \COMMENT{Average pooling across frames}
    \STATE // Compute text-video similarities
    \FOR{$t$ in \texttt{templates}}
        \STATE $z_t \gets \texttt{normalize}(G_t(t))$
        \STATE $s \gets \texttt{logit\_scale} \times (v \cdot z_t^\top)$
        \STATE \texttt{append} $s$ to \texttt{similarities}
    \ENDFOR
    \STATE $s_{avg} \gets \texttt{mean(similarities)}$
    \STATE $p \gets \texttt{softmax}(s_{avg} / \tau_c)$
\ENDFOR
\ENSURE \texttt{Class probabilities} $p$
\end{algorithmic}
\end{algorithm}

\begin{table}[hb]
\footnotesize
\centering
\renewcommand{\arraystretch}{1.1} 
\begin{tabular}{p{0.4\textwidth}}
\hline
\rowcolor{pink!30} \textbf{Scene and Temporal Context} \\
\hline
\textit{In this scene, something is \{\}.} \\
\textit{In this video, \{\} is happening.} \\
\textit{At this moment, the action being performed is \{\}.} \\
\textit{During this scene, something is \{\}.} \\
\textit{This video captures how something is \{\}.} \\
\textit{In this context, the action being shown is \{\}.} \\
\textit{This scene highlights the action of \{\}.} \\
\textit{The action happening right now is \{\}.} \\
\textit{Something is occurring in this scene: \{\}.} \\
\textit{During this time, \{\} is taking place.} \\
\hline
\rowcolor{blue!20} \textbf{Human-Centric Actions} \\
\hline
\textit{Look, the human is \{\}.} \\
\textit{A human can be observed performing \{\}.} \\
\textit{A human being is seen \{\} in this video.} \\
\textit{The individual is currently \{\}.} \\
\textit{This clip shows a person engaged in \{\}.} \\
\textit{This is a human performing \{\}.} \\
\textit{The person in the video is actively \{\}.} \\
\textit{In this video, a human is doing \{\}.} \\
\textit{The human action being performed is \{\}.} \\
\textit{Here, a person is captured \{\}.} \\
\hline
\rowcolor{green!20} \textbf{Questions} \\
\hline
\textit{What is happening in this video: \{\}?} \\
\textit{Can you tell what the action \{\} is?} \\
\textit{What activity is being demonstrated: \{\}?} \\
\textit{What can you observe in this scene: \{\}?} \\
\textit{Can you recognize the action \{\}?} \\
\textit{What is the person doing: \{\}?} \\
\textit{What action is depicted in this video: \{\}?} \\
\textit{Do you see what is happening here: \{\}?} \\
\textit{What movement is being performed: \{\}?} \\
\textit{Can you identify the activity \{\}?} \\
\hline
\rowcolor{orange!20} \textbf{Non-Human-Centric Prompts} \\
\hline
\textit{The creature is performing \{\}.} \\
\textit{This video shows a being engaged in \{\}.} \\
\textit{The action being done by the creature is \{\}.} \\
\textit{A creature is observed performing \{\}.} \\
\textit{This clip illustrates how a creature is \{\}.} \\
\hline
\end{tabular}
\caption{Categorized prompts for CLIP Zero-Shot classification, where the action class name will replace \{\}.}
\label{table:categorized_prompts}
\end{table}

\section{Additional Implementation Details}
\label{sec:rationale}
\textbf{Network Architecture.} As mentioned in Section \ref{Co-STAR} of the main paper, we employ identical network architectures for both teacher and student models, each consisting of a CLIP visual encoder followed by a Temporal Relation Network (TRN). Our TRN implementation follows Zhou \textit{et al.} \cite{zhou2018temporal}, maintaining their bottleneck dimension of 256, but increasing the \texttt{subsample\_num} from 3 to 9 to sample more frame relation combinations at each temporal scale. We use dropout with probability 0.5 after each ReLU activation for regularization. During all experiments, we keep the CLIP encoder frozen and only train the TRN module to prevent overfitting and maintain the encoder's robust visual representations.

\noindent\textbf{ CLIP Zero-Shot Video Classification.} Algorithm 1 shows the details of our zero-shot learning procedure for video classification. Similar to DALL-V \cite{zara2023unreasonable}, we utilize CLIP's vision-language alignment by encoding video frames and text templates, applying normalization, and computing scaled similarities between these modalities to obtain class probabilities. However, while DALL-V randomly selects one text template per sample \cite{zara2023unreasonable}, we leverage all text templates to compute the similarity scores, enabling more robust predictions through ensemble averaging.

Table \ref{table:categorized_prompts} lists the 35 text templates (prompts) used in our zero-shot classification, grouped into four categories. These prompts address various scenarios, including contextual descriptions, human and non-human actions, and questions about activities. This diversity enhances the alignment between video features and text templates to improve classification performance.

\section{Collabrative pseudo-labeling with MatchOrConf} 
As per in Section \ref{Co-STAR} of the main paper, we employ the MatchOrConf \cite{zhang2023rethinking} scheme to determine the collaborative pseudo-label $\tilde{y}$. 

Let $\mathcal{M}$ represent the $\arg\max$ operation, and $C_s$, $C_c$ represent the confidence scores from the teacher, and CLIP, respectively. Then, if $\xi = \{\mathcal{M}(p_{\theta_{s}}) == \mathcal{M}(p_{\theta_{c}})\}$, $\tilde{y}$ is assigned as: 
\begin{equation}
\footnotesize
\label{eq:matchconf}
\tilde{y} =
\begin{cases}
\mathcal{M}(p_{\theta_{s}}) & \text{if } \xi, \\
\mathcal{M}(p_{\theta_{s}}) & \text{if } !\xi \text{ and } C_s \geq \psi_s \text{ and } C_c < \psi_c, \\
\mathcal{M}(p_{\theta_{c}}) & \text{if } !\xi \text{ and } C_c \geq \psi_c \text{ and } C_s < \psi_s, \\
\mathcal{M}(p_{\theta_{s}}) & \text{if } !\xi \text{ and } C_s \geq \psi_s \text{ and } C_c \geq \psi_c \text{ and } C_s > C_c, \\
\mathcal{M}(p_{\theta_{c}}) & \text{if } !\xi \text{ and } C_s \geq \psi_s \text{ and } C_c \geq \psi_c \text{ and } C_c \geq C_s, \\
-1 & \text{otherwise.}
\end{cases}
\end{equation}

where the parameters $\psi_s$ and $\psi_c$ are the minimum confidence thresholds for teacher and CLIP respectively. Following \cite{reddy2024unsupervised}, we choose both $\psi_s$ and $\psi_c$ as 0.1.


\section{Detailed Formulations of Pace Functions}
In Section \ref{ablation_studies} of the main paper, we evaluate four different pace functions for our curriculum learning. The Exponential Pace Function is defined in Eq. \ref{eq-wCL}, and here we define the rest for completion.

Previous definitions in the paper were as follows: Let $r$ be the reliability score, $e' = e / E$ be the fraction of completed training (where $e$ is the current epoch and $E$ is the maximum number of epochs), and $w$ be the output weight. For all functions, $w_{\text{min}} = 0$ and $w_{\text{max}} = 1$ define the allowed range of weights. 
The parameter $\beta = 0.6$ is used across all functions, except for the stepwise Pace Function.

\noindent\textbf{Linear Pace Function.} This function provides a constant rate of increase throughout training:
\begin{equation}
w = \text{clamp}\left[r \cdot \left(1 + \beta \cdot e'\right), w_{\text{min}}, w_{\text{max}}\right]
\end{equation}

\noindent\textbf{Sigmoid Pace Function.} This function offers a smooth, S-shaped transition:
\begin{equation}
w = \text{clamp}\left[r_{\text{eqv}}+ \frac{1}{1 + \exp\left(-\beta \cdot \left(12e' - 6\right)\right)}, w_{\text{min}}, w_{\text{max}}\right]
\end{equation}
where $r_{\text{eqv}} = r(w_{\text{max}} - r)$ and the fraction of completed training scaled to $[-6, 6]$ for optimal sigmoid spread.

\noindent\textbf{Stepwise Pace Function.} This function provides discrete transitions in sample weights:
\begin{equation}
w = \text{clamp}\left[r + \left(\lfloor n \cdot e' \rfloor \cdot \frac{w_{\text{max}} - r}{n}\right), w_{\text{min}}, w_{\text{max}}\right]
\end{equation}
where $n = 4$, as suggested in \cite{wu2020curricula}, defines the number of discrete steps for weight updates.

\section{Impact of History Buffer Size} 
\begin{figure}[h]
\centerline{\includegraphics[scale=0.42]{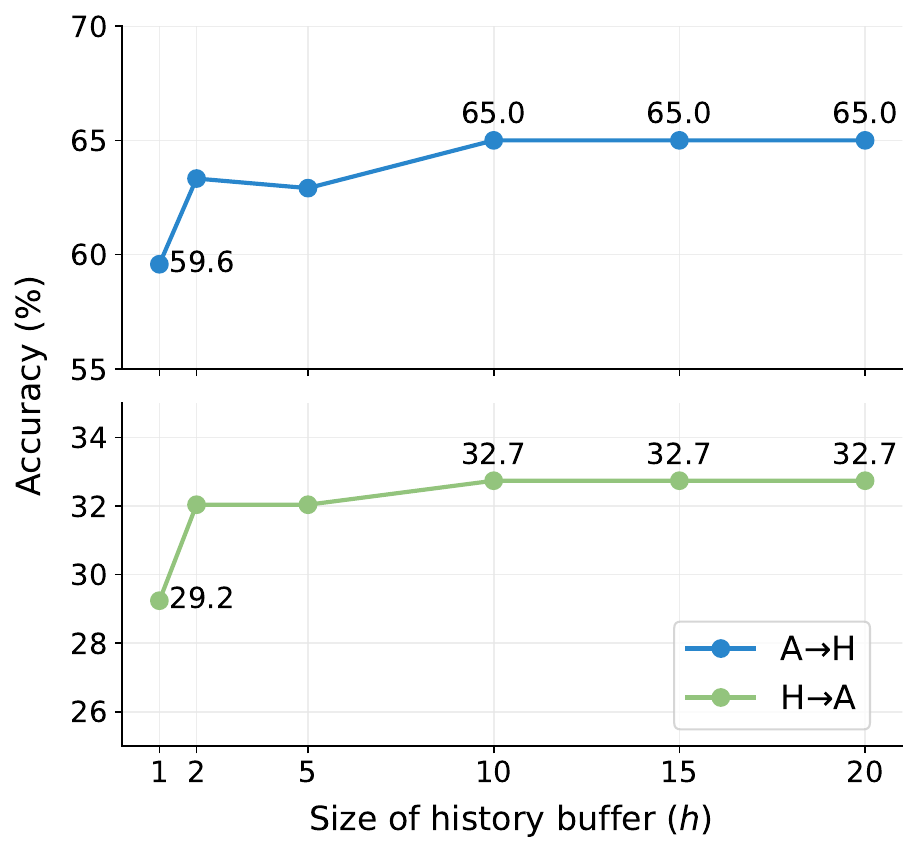}}
\caption{Impact of history buffer size $h$ on adaptation performance. The buffer  stores past predictions of the student model to assess prediction stability and stabilizes around $h=10$.}

\label{Fig:TQ-ablation} 
\end{figure}

{To evaluate the model's stability, we examine the impact of varying the history buffer size $k$, which stores historical predictions.
As shown in Fig.~\ref{Fig:TQ-ablation}, increasing the buffer size initially improves performance for both adaptation directions. The accuracy stabilizes after $k=10$, reaching 65.0\% for A$\rightarrow$H and 32.7\% for H$\rightarrow$A, suggesting that maintaining the last 10 predictions provides sufficient information for our framework.} 
Further increasing the buffer size does not show additional benefits. 

\section{Hyperparameters} The optimized hyperparameters for our curriculum learning and ACR components, determined empirically, are shown in Table~\ref{tab:hyperparameters}. To evaluate their robustness, we also tested the sensitivity of these parameters across different ranges, as detailed in Table~\ref{tab:param_ranges} for curriculum learning and Table~\ref{tab:parameter_ranges_2} for ACR. 

For these evaluations, we chose the A → H (ARID → HMDB51) setting due to its larger domain gap and shorter training time compared to other scenarios. 

\begin{table}[h!]
\small
\centering
\begin{tabular}{l|c|c}
\hline

\textbf{Component}     & \textbf{Parameter} & \textbf{Optimal Value} \\ \hline
\multirow{2}{*}{Curriculum Learning} & $\alpha$ (Eq. \ref{eq:r})           & 0.5            \\ 
                                     & $\beta$ (Eq. \ref{eq-wCL})            & 0.6            \\ \hline
\multirow{4}{*}{ACR}                 & $\eta$ (Eq. \ref{eq:ACR})            & 6              \\ 
                                     & $\rho$ (Eq. \ref{eq:pace-exp})            & 0.25           \\ 
                                     & $\sigma$ (Eq. \ref{eq:w-update})          & 0.05           \\ 
                                     & $\gamma$ (Eq. \ref{eq:w-update})          & MaxConf        \\ 
                                     & $\lambda$ (Eq. \ref{eq:if})          & 0.2        \\ \hline
                                     
\end{tabular}
\caption{Hyperparameters for Curriculum Learning and ACR. }
\label{tab:hyperparameters}
\end{table}


\begin{table}[h!]
\small
\centering
\begin{tabular}{c|c|c}
\hline
\textbf{Parameter} & \textbf{Range} & \textbf{A$\rightarrow$H} \\ \hline
\multirow{3}{*}{$\alpha$} & 0.25           & 57.0                     \\ 
                          & \bf{0.5}            & \bf{60.8}                   \\ 
                          & 0.75           & 59.6                   \\ \hline
\multirow{4}{*}{$\beta$}  & 0.2            & 57.1                    \\ 
                          & 0.4            & 60.0                     \\ 
                          & \bf{0.6}            & \bf{60.8}                   \\ 
                          & 0.8            &  60.5                         \\ \hline
\end{tabular}
\caption{Performance of different parameter ranges for Curriculum Learning. Note that the ACR component was excluded in this analysis to better evaluate the curriculum hyperparameters.}
\label{tab:param_ranges}
\end{table}

\begin{table}[h]
\small
\centering
\begin{tabular}{c|c|c}
\hline
\textbf{Parameter} & \textbf{Range} & \textbf{A$\rightarrow$H} \\ \hline
\multirow{4}{*}{$\eta$}  & 1            & 62.3                  \\ 
                          & 3            & 63.3                  \\ 
                          & \bf{6}            & \bf{65.0}                    \\ 
                          & 9            & 60.8                  \\ \hline
\multirow{4}{*}{$\rho$}  & 0.75         & 62.5                   \\ 
                          & 0.5          & 61.6                  \\ 
                          & \bf{0.25}         & \bf{65.0}                    \\ 
                          & 0.1          & 63.8                   \\ \hline
\multirow{3}{*}{$\sigma$} & 0.07          &  63.8                        \\ 
                          & \bf{0.05}         & \bf{65.0}                    \\ 
                          & 0.03         &    62.9                    \\ \hline
\multirow{4}{*}{$\lambda$} & 0.7        & 59.0                    \\ 
                          & 0.5          & 62.1                   \\ 
                          & \bf{0.2}         & \bf{65.0}                    \\ 
                          & 0.1          & 64.1                   \\ \hline
\end{tabular}
\caption{Performance of different parameter ranges for ACR.}
\label{tab:parameter_ranges_2}
\end{table}










To calculate $\gamma$ (confidence threshold, Eq. \ref{eq:w-update} of the main paper), we use the maximum confidence (MaxConf) of samples where the teacher and CLIP predictions agree, i.e.,

\begin{equation}
\gamma = \max_{i \in \mathcal{A}} \{C_s^i\}
\end{equation} 

where $\mathcal{A} = \{i \in \{1,...,X_t\} : \mathcal{M}(p_{\theta_s}(x_i^t)) = \mathcal{M}(p_{\theta_c}(x_i^t))\}$ represents the set of indices where teacher and CLIP predictions agree, $p_{\theta_s}(x_i^t)$ and $p_{\theta_c}(x_i^t)$ are the predicted probability distributions of the teacher and CLIP models for target sample $x_i^t$ respectively, and $C_s^i$ represents the confidence score of the teacher model for sample $i$.

\end{document}